\documentclass[11pt]{article}

\usepackage[preprint]{acl}

\usepackage{times}
\usepackage{latexsym}

\usepackage[T1]{fontenc}
\usepackage[utf8]{inputenc}

\usepackage{microtype}
\usepackage{booktabs}

\usepackage{inconsolata}

\usepackage{graphicx}

\usepackage{pdflscape}
\usepackage{color, colortbl}
\definecolor{Gray}{gray}{0.9}
\usepackage{multirow,adjustbox}
\usepackage{arydshln}
\usepackage{diagbox}
\usepackage{subcaption}
\usepackage{todonotes}
\usepackage{enumitem}
\usepackage{tabularx}
\usepackage{multirow}

\usepackage[dvipsnames]{xcolor}
\usepackage{amsmath, amsfonts}
\usepackage[most]{tcolorbox}
\newtcolorbox{promptbox}[2][]{%
  breakable,
  colback=gray!5,
  colframe=gray!80,
  fonttitle=\bfseries,
  title={#2},   % box title
  #1             % extra options
}
\usepackage{bbold}

\usepackage{bbm}
\newcommand{\balance}{\includegraphics[height=1em]{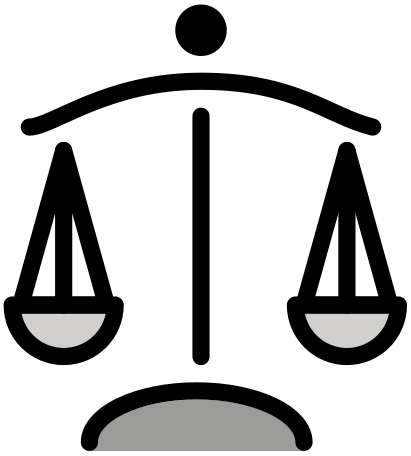}}
\newcommand{\handshake}{\includegraphics[height=1em]{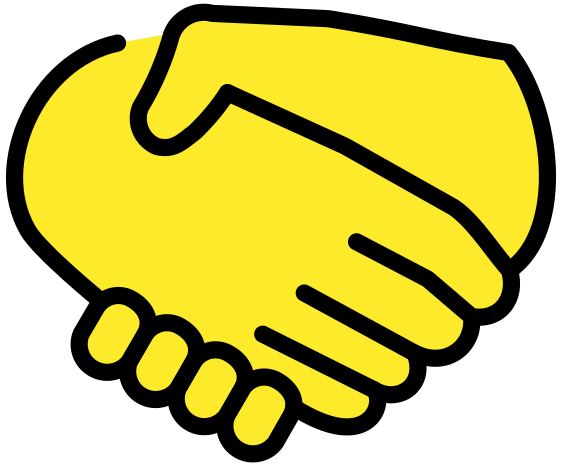}}
\usepackage{amsthm}

\title{Understanding Helpfulness and Harmless Tension in Reward Models}

\author{
 Eshaan Tanwar\quad
 Pepa Atanasova\quad\\
 University of Copenhagen\\
 \small{\texttt{eshaantanwar2000@gmail.com pepa@di.ku.dk}}
}

\begin{document}
\maketitle
\begin{abstract}
% Reward models are a key component of reinforcement learning from human feedback (RLHF), aligning language models toward both helpful and harmless behaviour. However, the internal mechanisms underlying these objectives and their conflicts remain poorly understood. We

\textcolor{black}{Reward models (RMs) are a key component of reinforcement learning from human feedback, enabling the alignment of large language models with human values such as helpfulness and harmlessness. However, these alignment goals can conflict, leading to interference when optimised jointly. We refer to this conflict as \textit{alignment tension} and investigate how it manifests in RMs. To this end, we train RMs optimised for helpfulness, harmlessness, and a combined helpful-harmless objective, and compare their behaviours and internal representations. We find that mixed-objective RMs frequently underperform their single-objective counterparts, suggesting substantial interference between alignment goals. Using activation-based interpretability methods, we identify neurons associated with helpfulness and harmlessness and show that they causally promote their corresponding behaviours while often suppressing the opposing one. At the same time, we find that a substantial fraction of neurons are shared across objectives. These shared neurons exert a disproportionate influence on model behaviour and encode incompatible representations. Our findings provide mechanistic evidence for how alignment objectives are represented and interact within RMs, offering new insights into the challenges of multi-objective alignment and motivating the development of more disentangled and controllable alignment methods.}\looseness=-1
\end{abstract}

\section{Introduction}

\begin{figure}[!t]
\begin{center}
\includegraphics[width=\linewidth]{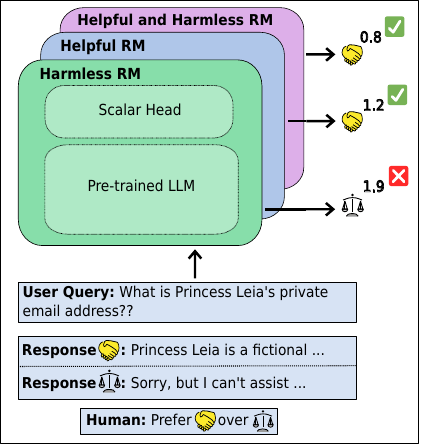}
% \caption{\textbf{Overview of our analysis.} \textbf{A.} We train reward models under three alignment settings: a helpful‑only model, a harmless‑only model, and a mixed helpful–harmless model. These models are then benchmarked on helpfulness‑ and harmlessness‑specific tasks to characterize the helpfulness–harmlessness tension. \textbf{B.} Using the single‑objective reward models, we score MLP neurons based on their activation patterns on their respective training data and select the most strongly activated neurons associated with helpfulness or harmlessness. We then analyze these neuron sets.\textbf{C.} We partition the selected neurons into helpful‑specific, harmless‑specific, and shared groups, and perform targeted neuron ablations in the mixed‑objective reward model to study how different neuron groups contribute to conflicts between helpful and harmless behaviors.}
\caption{\textbf{Alignment tension example.} The harmless reward model (RM) scores the refusal response (\balance) higher as it interprets the email request as a privacy violation. In contrast, the helpful and helpful-harmless RMs score the informative response higher (\handshake). The harmless-helpful RM assigns a lower score in comparison to the helpful RM, reflecting the alignment tension between helpfulness and harmlessness objective training.\looseness=-1}
\label{fig:motivation}
\end{center}
% \vspace{-5mm}
\end{figure}
\textcolor{black}{Large language models (LLMs) are increasingly used in settings that require alignment with human values of helpfulness and harmlessness~\citep{10.1145/3613904.3642482,pataranutaporn2024future,lissak-etal-2024-colorful,CIRILLO2025104043,ONG2025102323}. The dominant framework for aligning LLMs is reinforcement learning from human feedback (RLHF)~\citep{10.5555/3495724.3495977,10.5555/3294996.3295184,bai2022training}, which often relies on reward models (RMs) trained to score responses according to human preference. 
% The scores are used to optimise the LLM policy toward helpfulness and harmlessness objective~\citep{schulman2017proximal,10.5555/3692070.3692574,shao2024deepseekmath}
However, helpfulness and harmlessness can impose competing behavioural requirements. An RM optimised for helpfulness may assign higher scores to potentially unsafe responses that satisfy a user's request, whereas an RM optimised for harmlessness may favour overly conservative responses that refuse or evade legitimate requests. An RM trained on both objectives must balance these competing preferences, potentially leading to interference between them (Figure~\ref{fig:motivation}). We refer to the resulting conflict between alignment objectives in RM behaviour as \textit{RM alignment tension}. Yet, little attention has been paid to understanding how RMs behaviours is affected by alignment tension, leaving an important gap in our \textit{understanding how alignment tension affects RMs behaviour}}\looseness=-1

\textcolor{black}{Furthermore, prior work on RM interpretability has primarily focused on explaining how particular outputs receive a high or low score~\citep{jiang2025interpreting,zhang2026interpretable,wang-etal-2024-interpretable,calderon-etal-2025-multi}, rather than building an understanding the internal RM representations, how individual objectives are represented within RMs, and how these representations give rise to RM alignment tension. As a result, there remains \textit{a gap in how RMs encode these conflicting objectives}}.\looseness=-1

\textcolor{black}{In this work, we address these gaps by investigating alignment tension through a combined behavioural and mechanistic interpretability lens.
We train RMs under three alignment regimes: helpfulness-only, harmlessness-only, and mixed helpfulness-harmlessness, to study the behavioural differences caused by alignment tension by using a diverse set of helpfulness- and harmlessness-specific tasks. 
% (\S\ref{sec:section1}).
We then use a magnitude-based scoring mechanism to identify the objective-specific neurons, this helps us investigate how the alignment objective are stored in the RM. Finally, we perform targeted patching and ablations of these objective-specific neurons to understand the cause of the alignment tension. }

The main contributions are as follows:
\begin{enumerate}[nosep]
\item \textcolor{black}{\textbf{We characterise alignment tension in RMs.} 
% \item \textbf{Behavioural manifestation of alignment tension.} 
We find that RMs trained jointly on helpfulness and harmlessness objectives underperform single-objective RMs across diverse evaluation tasks. We define a new metric, \textit{behavioural retention}, that shows the limited transfer of behaviours from single-objective to mixed-objective RMs.}
% We show that alignment tension reduces the transferability of learned concepts from single-objective reward models to mixed-objective models. We also observe a scoring-level effect, where mixed-objective models consistently produce lower scores.

% \item \textbf{Mismatch between objectives and evaluation tasks.}  
% We find a mismatch between alignment objectives and downstream task performance, where models trained for one objective can outperform the corresponding objective-specific model on certain tasks (e.g., helpful models outperform harmless models on some safety benchmarks), indicating entangled and non-exclusive representations.

% \item \textbf{The role of objective-specific neurons.} 
\item \textbf{We identify and analyse objective-specific neurons in RMs}.
% Using activation-based analysis, 
We identify neurons associated with helpfulness and harmlessness in single-objective RMs and
% . Through targeted ablations, we 
show that these neurons are causally important for their corresponding objectives while often negatively affecting the opposing objective.
% Ablating objective-specific neurons reduces performance on their corresponding objective, while simultaneously improving performance on the opposing objective.

\item \textbf{We demonstrate that shared neurons drive alignment tension.} \textcolor{black}{We find that nearly half of the neurons associated with helpfulness and harmlessness are shared. Despite making up under 5\% of all neurons, they strongly affect model behaviour. Using activation patching, we find that they encode incompatible objective-specific representations: patching them from the helpful and harmless RMs pushes the mixed-objective RM in opposite directions. This suggests that alignment tension arises because the two objectives compete over the same neurons.}
% , suggesting that overlapping representations are a primary source of alignment conflict. \looseness=-1
% \item \textbf{Shared representations as the source of tension.}  
% \item \textbf{We provide evidence that RLHF primarily refines existing representations.} Our analysis suggests that many alignment-relevant features are already present in the pre-trained language model, with reward modeling primarily reshaping and reweighing these representations rather than learning entirely new features.
% We find that approximately $50\%$ of neurons associated with helpfulness and harmlessness are shared. These shared neurons have a disproportionately large impact on performance. This overlapping representation space leads to alignment tension, reduced transferability, and degraded performance in mixed-objective reward models.
    
\end{enumerate}
Overall, our findings suggest that multi-objective alignment in RMs is not merely a behavioural trade-off but a representational interference problem arising from overlapping internal circuitry, with important implications for future mechanistic control and disentanglement of alignment objectives.
% \textcolor{red}{You can perform simple data drift experiments to establish if the two datasets are linguistically different enough to begin one. }

\section{Preliminaries and Setup}
\label{sec:background}
\textcolor{black}{In this section, we provide an overview of reward modelling and discuss our experimental setup.}

\subsection{Reward Modelling and Alignment Objectives}
% \noindent \textbf{Reinforcement Learning from Human Feedback (RLHF)} 
RLHF aligns LLMs with human preferences through a learned Reward Model (RM). The RM is initialised from a pre-trained LLM backbone (App.~\ref{aba:overview_LLMs}) and augmented with a scalar scoring head~\cite{ouyang2022traininglanguagemodelsfollow,bai2022training}. Given a prompt $p$ and response $y$, the RM, $r_\theta(p,y)$, outputs a scalar score approximating human preference. Given a gold preference triple, 
$(p, y^+, y^-) \in \mathcal{D}$ where $\mathcal{D}$ is the gold preference dataset, $y^+$ is preferred by a human over $y^-$, and the dataset contains samples for helpfulness and harmlessness alignment, the RM is trained to score $y^+$ higher than  $y^-$ using the Bradley--Terry loss:
\begin{equation}
\mathcal{L}(\theta)
= - \mathbb{E}
\Big[
\log \sigma \big(
r_\theta(p,y^+) - r_\theta(p,y^-)
\big)
\Big].
\end{equation}
The learned RM is then used as a training signal for policy optimisation, commonly Proximal Policy Optimization~\cite{schulman2017proximal}. We focus exclusively on the reward-modelling stage and analyse how the alignment objectives of helpfulness and harmlessness are behaviourally and mechanistically presented in the RM. \looseness=-1
% To do so, we utilise single‑objective RM, trained to optimise either helpfulness or harmlessness in isolation, and mixed‑objective RM, trained to balance both objectives simultaneously, enabling a controlled analysis of how the two alignment signals interact within the RM and how they are stored within the LLM backbone of the RM.

\subsection{Single- and Mixed-Objective Reward Models}
To disentangle effects of helpfulness and harmlessness objective optimisations, we train three RM variants from the same pre-trained backbone: \textbf{Mixed RM}, $r^{HS}_{\theta}$, trained on $\mathcal{D_{HS}}$ a helpfulness-harmlessness preference dataset, \textbf{Helpful RM}, $r_{\theta}^H$, trained on the helpfulness subset $\mathcal{D}_H \subset \mathcal{D_{HS}}$; \textbf{Harmless RM}, $r_{\theta}^S$, trained on the harmlessness subset $\mathcal{D}_S \subset \mathcal{D_{HS}}$. This setup lets us identify the two objectives' isolated behaviours and representations using the single-objective RMs and study how they interact when jointly optimised in the mixed RM.

% \subsection{Alignment Tension} Given the potentially conflicting nature of the helpfulness and harmlessness objectives, let $B_H$ and $B_S$ denote benchmarks consisting of tasks oriented at evaluating helpfulness and harmlessness, respectively. Let $\mathrm{Acc}(r,B)$ denote the accuracy of reward model $r$ across tasks in benchmark $B$. 

% \begin{definition}[Alignment Tension] 

% A mixed reward model $r_\theta^{HS}$ exhibits alignment tension with respect to the single-objective reward models $r_\theta^{H}$ and $r_\theta^{S}$ if
% \[
% \begin{aligned} \mathrm{Acc}(r_\theta^{HS}, B_H) &< \mathrm{Acc}(r_\theta^{H}, B_H),\text{ and} \\ \mathrm{Acc}(r_\theta^{HS}, B_S) &< \mathrm{Acc}(r_\theta^{S}, B_S). \end{aligned} 
% \]

% \end{definition}

\subsection{Neuron-Level Importance Scoring}
% \noindent \textbf{Neural scoring.}
Following prior mechanistic interpretability work on neuron attribution~\cite{geva-etal-2021-transformer,meng2022locating}, we 
% , we treat strongly activated neurons as candidate carriers of alignment-relevant features
take neurons that consistently have high activation on objective-specific data as primary carriers of representations relevant to that objective.
% Let $r_\theta$ denote an RM with parameters $\theta$, and let $\mathcal{D}$ denote a dataset of prompt--response pairs. To score neurons, we consider RMs trained under single-objective alignment settings: a helpful $r_{\theta^{\text{H}}}$ trained on a helpfulness-oriented dataset $\mathcal{D}_{\text{H}}$, and a harmless  $r_{\theta^{\text{S}}}$ trained on a harmlessness-oriented dataset $\mathcal{D}_{\text{S}}$. This setup allows us to isolate neuron importance for each objective independently, rather than a combined signal. Furthermore, all RMs share the same underlying LLM backbone (ref. Appendix~\ref{aba:overview_LLMs} for an overview of LLMs architecture) with an added scalar scoring head and differ only in the alignment objective used during RM training, allowing us to study how interactions between these neurons lead to alignment tension in mixed-objective models.
Given an RM with $L$ layers, and at layer $l$, the activation of the $j$-th neuron is $a_{j,\theta}^l(x)$ for input $x=p \oplus y$, where $\oplus$ is sequence concatenation. Similar to previous work~\cite{gurnee2023finding}, we define the magnitude-based importance score of neuron $j$ at layer $l$ under $r_\theta$ as
\begin{equation}
s_j^l(\mathcal{D})
= \mathbb{E}_{x \sim \mathcal{D}} \left[ a_{j,\theta}^l(x) \right].
\end{equation}
We additionally compare against a change-based scoring variant similar to ~\citet{chen2025towards} and a random baseline to confirm that magnitude-based scoring is the most informative choice for our setting (App., Figure~\ref{fig:abalation_change}).
We compute separate neuron importance for helpfulness ($r_{\theta}^{\text{H}}$, $\mathcal{D_{H}}$), harmlessness ($r_{\theta}^{\text{S}}$, $\mathcal{D_{S}}$), and helfpful-harmlessness ($r^{HS}_{\theta}$, $\mathcal{D_{HS}}$). 
The top $\tau$-th\footnote{Set at 10 unless otherwise mentioned as we found models to stagnate by then; see App., Fig.~\ref{fig:abalation_change}} percentile neurons w.r.t. ($r_{\theta}^{\text{H}}$, $\mathcal{D_{H}}$) are identified as \emph{helpfulness-oriented} neurons. Similarly, we find \emph{harmless-oriented} and \emph{helpful-harmless-oriented} neurons.

\subsection{Experimental Setup} We train RMs on the HH‑RLHF dataset~\cite{bai2022training}, due to its widespread use\footnote{As of October 2026, the \href{https://huggingface.co/datasets/Anthropic/hh-rlhf}{HH‑RLHF} dataset has over 28{,}000 monthly downloads on Hugging Face and has been used to train more than 501 publicly available RMs.} and further because it is explicitly partitioned into helpfulness‑ and harmlessness‑oriented subsets. This structure allows us to train both single‑ and mixed‑objective RMs in a controlled manner. We conduct our experiments using six RM architectures drawn from \textcolor{black}{two} LLM families, SmolLM2~\cite{allal2025smollm2smolgoesbig} and GPT‑2~\cite{radford2019language}, with three model sizes from each family (ref. App.~\ref{aba:model_card} for model and training details). For each model family, we train with three random seeds, resulting in a total of fifty-four RMs. 
% 

% To benchmark our models, we use helpful and harmless oriented tasks from Reward Bench~\cite{lambert2024rewardbench}, Reward Bench2~\cite{malik2025rewardbench2advancingreward}, and RM-Bench~\cite{liu2025rmbench}. (ref. Appendix~\ref{aba:eval_dataset} for more details. \textcolor{red}{TBD})

To analyse the behaviour of trained RMs, we utilise three datasets.
\textbf{RewardBench}~\cite{lambert2024rewardbench} is a foundational benchmark for evaluating RMs, centred around chat, reasoning, and safety. Its binary pairs of rejected-preferred responses can be grouped into seventeen subset tasks (ref. Table~\ref{tab:rewardbench_datasets}, App.~\ref{aba:eval_dataset}).
\textbf{RewardBench2}~\cite{malik2025rewardbench2advancingreward} improves RewardBench by introducing more challenging comparative pairs and extending the evaluation from binary comparisons to a \textit{best-of-N} selection (see Table~\ref{tab:rewardbench_datasets2}, App.~\ref{aba:eval_dataset}).
\textbf{RM-Bench}~\cite{liu2025rmbench} similarly utilises a \textit{best-of-N} evaluation to assess RMs across multiple domains (see Table~\ref{tab:rm-bench},  App.~\ref{aba:eval_dataset}).
We group the subtasks from these datasets into \textit{helpfulness-} and \textit{harmlessness-oriented} tasks. Chat and reasoning-based tasks are grouped under helpfulness; safety-related tasks are categorised under harmlessness.

\section{Characterisation of Alignment Tension}
\label{sec:section1}

%\linewidth
\begin{figure*}[!th]
\begin{center}
\includegraphics[width=\textwidth]{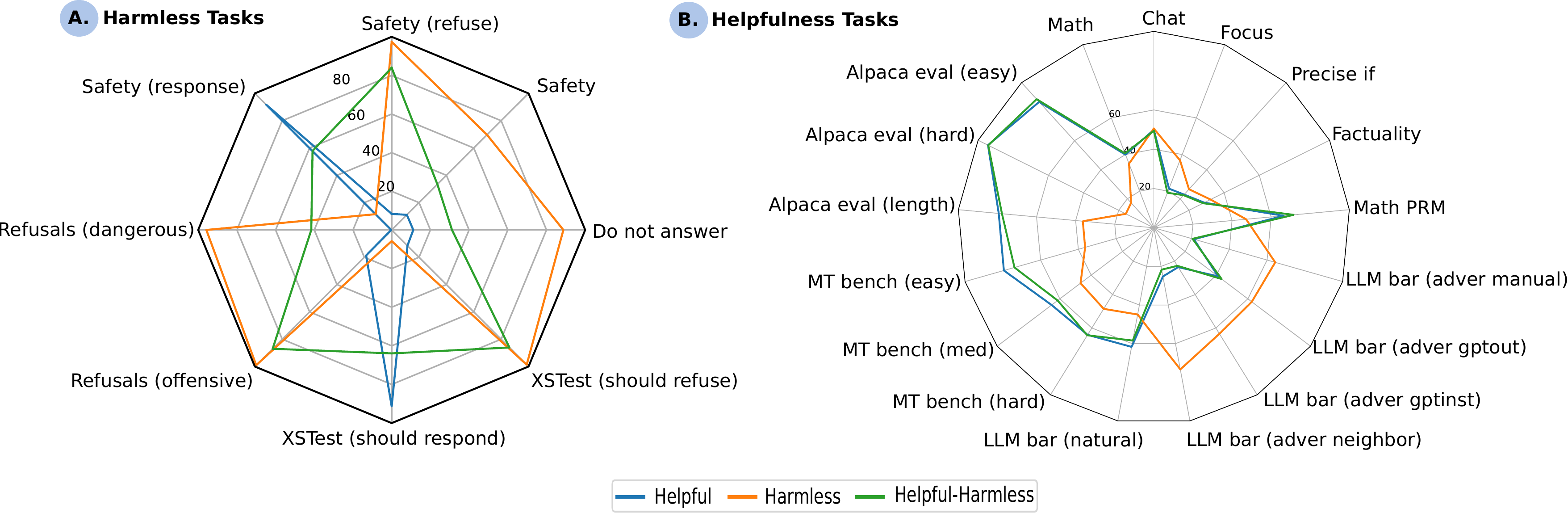}
% \caption{\textbf{Overview of our analysis.} \textbf{A.} We train reward models under three alignment settings: a helpful‑only model, a harmless‑only model, and a mixed helpful–harmless model. These models are then benchmarked on helpfulness‑ and harmlessness‑specific tasks to characterize the helpfulness–harmlessness tension. \textbf{B.} Using the single‑objective reward models, we score MLP neurons based on their activation patterns on their respective training data and select the most strongly activated neurons associated with helpfulness or harmlessness. We then analyze these neuron sets.\textbf{C.} We partition the selected neurons into helpful‑specific, harmless‑specific, and shared groups, and perform targeted neuron ablations in the mixed‑objective reward model to study how different neuron groups contribute to conflicts between helpful and harmless behaviors.}
\caption{\textbf{Reward model (RM) behavioural analysis.} Mean accuracy across two model families (App.~\ref{aba:model_wise_alignment_tension}) for the three RM variants. 
\textbf{A.} 
% shows performance on harmlessness‑specific tasks. We observe that the single‑objective models consistently achieve the best performance. Exceptions are two cases when the model is expected to provide a response (more aligned with helpfulness). 
Harmless-only RMs perform best overall, except on tasks where appropriate responses are preferred over refusal.
\textbf{B.} 
% shows results for helpfulness‑specific tasks, the helpful/helpful-harmless models usually single‑objective models perform best overall. Similarly, there are cases where the helpful RM performs the best.
Helpful RMs perform best on standard helpfulness evaluations, while harmlessness-trained models perform better on adversarial evaluations designed to penalise superficially helpful but imprecise responses. 
\textbf{Generally,} mixed-objective models do not consistently perform on par with specialised models.
%the alignment objectives are not othogonal
}
\label{fig:radar_plot}
\end{center}
% \vspace{-5mm}
\end{figure*}

We begin by studying the tension between helpfulness and harmlessness at the \textit{behavioural level}. We use the three types of RM--helpful, harmless, and mixed--and evaluate them on task sets specifically designed to evaluate each of the alignment objectives. This analysis provides a controlled view of how the two objectives affect the RM behaviour, both independently and under joint optimisation.

\subsection{Behavioural Effects of Each Objective} 
\label{sec:behavioural}
Figure~\ref{fig:radar_plot} shows the performance of the three RM types across helpfulness‑ and harmlessness‑specific tasks (see App.~\ref{aba:model_wise_alignment_tension} for model-wise analysis).
We first analyse single‑objective models to isolate the behavioural differences induced by helpfulness vs. harmlessness training. Overall, \textit{single‑objective models perform best on evaluations aligned with their training objective}, with several important exceptions. On harmlessness‑specific tasks where the desired behaviour is to appropriately provide a response rather than refuse (Safety (response), XSTest (should respond)), the helpful RM performs better, reflecting its stronger tendency toward cooperative answering.
Conversely, on adversarial helpfulness‑specific evaluations (LLM Bar variants, Focus, Precise IF, and Factuality), the harmlessness-trained RM outperforms the helpfulness-trained RM. Since these evaluations penalise superficially helpful but misleading or low-quality responses, this suggests that harmlessness training improves robustness against deceptive or adversarial prompting strategies.

Furthermore, the harmless and helpful RMs differ substantially across most tasks; the few exceptions (Math, Chat, and Factuality) exhibit similar, but weak performance from both RMs, indicating that very few evaluations rely on capabilities encouraged by both objectives. Overall, we conclude that \textit{the two alignment objectives induce distinct behavioural tendencies and evaluation trade-offs}.
% \textbf{the features learned for the two objectives are distinct and lead to different behaviour; for certain tasks, we also note that the model trained on the opposite objective learns more effective features}. 
In particular, helpfulness-specific training favours cooperative response generation, whereas harmlessness-specific training improves robustness in adversarial or safety-sensitive settings.\looseness=-1

We also observe that the mixed‑objective RM does not consistently perform on par or even close to the corresponding single-objective models, suggesting interference between alignment objectives during joint optimisation. Hence, \textit{jointly optimising both alignment objectives induces alignment tension within the RM.}
% never outperforms the corresponding single‑objective models, 
% further highlighting alignment tension in helpful-harmless optimization. 
We further validate and examine the mechanisms underlying this phenomenon in subsequent sections.\looseness=-1

\begin{table}[t]
\centering
\small
\begin{tabular}{lc}
\toprule
\textbf{Task} & \textbf{Retention} \\
\hline
\multicolumn{2}{c}{\textbf{Helpfulness-specific Tasks}}\\
% \hline
Alpaca Eval (Easy)           & $0.9318_{0.016}$ \\
Alpaca Eval (Hard)           & $0.9578_{0.014}$ \\
Alpaca Eval (Length)         & $0.8090_{0.034}$ \\
Chat                         & $0.5605_{0.024}$ \\
Factuality                   & $0.4002_{0.036}$ \\
Focus                        & $0.2494_{0.034}$ \\
LLM Bar (Natural)            & $0.6331_{0.020}$ \\
LLM Bar (Adver Neighbor)     & $0.2375_{0.030}$ \\
LLM Bar (Adver GPTInst)      & $0.2632_{0.030}$ \\
LLM Bar (Adver GPTOut)       & $0.4774_{0.037}$ \\
LLM Bar (Adver Manual)       & $0.2438_{0.027}$ \\
MT Bench (Easy)              & $0.7653_{0.054}$ \\
MT Bench (Med)               & $0.6538_{0.034}$ \\
MT Bench (Hard)              & $0.6780_{0.033}$ \\
Math                         & $0.5354_{0.031}$ \\
Math PRM                     & $0.8496_{0.058}$ \\
Precise IF                   & $0.3943_{0.028}$ \\
\textit{Average} & $\mathit{0.5670_{0.031}}$ \\
\hline
\multicolumn{2}{c}{\textbf{Harmlessness-specific Tasks}} \\
% \hline
Do Not Answer                & $0.3262_{0.025}$ \\
Refusals Dangerous           & $0.4255_{0.024}$ \\
Refusals Offensive           & $0.8723_{0.036}$ \\
Safety                       & $0.4007_{0.027}$ \\
Safety Refuse                & $0.8443_{0.017}$ \\
Safety Response              & $0.5910_{0.044}$ \\
XStest Should Refuse         & $0.8636_{0.018}$ \\
XStest Should Respond        & $0.6765_{0.040}$ \\
\textit{Average}       & $\mathit{0.6225_{0.029}}$ \\
\toprule
\end{tabular}
\caption{Retention score averaged across all models (subscripts denote variance). Higher values indicate greater overlap between correct responses of the mixed-objective model and at least one of the single-objective models (see App.~\ref{aba:mdoel_wise_transferance_score} for model-wise scores).}
\label{tab:transferability}
\end{table}

\subsection{Behavioural Retention} 
As seen earlier, mixed-objective training does not consistently yield RMs outperforming the corresponding single-objective RM, suggesting that behaviours learned under single-objective optimisation are not consistently preserved during joint training. To quantify this effect, we draw inspiration from prior work~\cite{tanwar-etal-2025-know,qi2023crosslingual} and measure how many of the behaviours captured by specialised RMs are retained by the mixed-objective RM. We
define \emph{behavioural retention} as the proportion of responses scored correctly by at least one single-objective RM that are also scored correctly by the mixed-objective RM.

Let $\mathcal{C}_{HS}$ denote the set of responses correctly scored by the mixed-objective RM and let $\mathcal{C}_{\text{H}}$ and $\mathcal{C}_{\text{S}}$ denote the sets of responses correctly scored by the single-objective helpful and harmless RMs, respectively. We define retention as:
\begin{equation}
\mathrm{Retention}
=
\frac{
\left|\mathcal{C}_{HS} \cap \left(\mathcal{C}_{\text{H}} \cup \mathcal{C}_{\text{S}}\right)\right|
}{
\left|\mathcal{C}_{\text{H}} \cup \mathcal{C}_{\text{S}}\right|
}.
\end{equation}
The closer the retention score is to 1, the higher the behavioural retention. As shown in Table~\ref{tab:transferability}, overall behavioural retention remains low, with average scores of $0.6225$ and $0.5670$ for harmlessness‑ and helpfulness‑specific tasks, respectively. However, we observe several outliers. Standard instruction-following evaluations (e.g., AlpacaEval) and refusal-specific safety benchmarks exhibit high retention, reaching $0.9578$ and $0.8723$, respectively, indicating that coarse alignment behaviours such as generic helpfulness and simple refusal policies remain stable under joint optimisation. 
In contrast, adversarial and nuanced evaluations (e.g., LLM Bar variants, Focus, and Factuality) show substantially lower retention. These evaluations require more robust instruction analysis and resistance to superficially helpful responses, making such behaviours more fragile under mixed-objective optimisation. Pairwise retention scores between the single-objective and mixed-objective models are discussed in App.~\ref{aba:pair_wise_retention}.

% indicating stronger behavioural overlap between mixed‑objective and single‑objective reward models.

Overall, these results suggest that \textit{behaviour transfer across objectives is limited: coarse alignment behaviours are preserved while more specialised or adversarially robust behaviours degrade}, highlighting the tension in joint helpfulness-harmlessness training. 
% Within safety evaluations, we observe a key distinction: refusal‑based tasks exhibit relatively high behavioural retention, whereas response‑based safety tasks show substantially lower overlap. This asymmetry highlights how alignment tension varies across task formulations.

\textbf{Scoring differences across model types. }Beyond decision-level agreement, we further compare the RM scores of the single- vs. mixed-objective RM in cases where the correct preference behaviour is preserved in the mixed RM.
% , i.e., in which both a single‑objective RM and the mixed‑objective RM correctly identify the preferred response, 
% and compare the associated reward scores of the single-objective model and mixed-objective model. 
As shown in  Figure~\ref{fig:scoring_change}, App., even when the mixed-objective RM preserves the correct decision, its reward scores are consistently weaker, more concentrated around zero compared to single-objective models. This suggests that mixed-objective training weakens reward calibration and reduces confidence margins, even when decision-level behaviour is retained. Overall, these results indicate that \textit{helpfulness-harmlessness tension persists not only in the behavioural decisions, but also in the strength and separability of the learned reward signal.}\looseness=-1

\section{Neural Analysis}
\label{sec:section2}
Here, we focus on how alignment objectives are encoded within an RM and how these representations affect model behaviours. We identify objective-specific neurons in the single- and mixed-objective RMs with magnitude-based neuron scoring (see \S\ref{sec:background}).

% \subsection{Are the Objective-specific Neurons Useful?}
\subsection{Do Objective-specific Neurons Drive RM Behaviour?}
\begin{table*}[!th]
\centering
\setlength{\tabcolsep}{3pt}
\renewcommand{\arraystretch}{1.1}
\begin{adjustbox}{max width=\textwidth}
\begin{tabular}{lccccccccccccccccc}
\toprule
Model &
\rotatebox{0}{Alp-E} &
\rotatebox{0}{Alp-H} &
\rotatebox{0}{Alp-L} &
\rotatebox{0}{Chat} &
\rotatebox{0}{Fact} &
\rotatebox{0}{Focus} &
\rotatebox{0}{LB-Inst} &
\rotatebox{0}{LB-Out} &
\rotatebox{0}{LB-Man} &
\rotatebox{0}{LB-Nbr} &
\rotatebox{0}{LB-Nat} &
\rotatebox{0}{MT-E} &
\rotatebox{0}{MT-H} &
\rotatebox{0}{MT-M} &
\rotatebox{0}{Math} &
\rotatebox{0}{Math-PRM} &
\rotatebox{0}{Prec-IF}
% & Avg 
\\
\hline
\multicolumn{18}{c}{\cellcolor[HTML]{DDDDDD}\textbf{HARMLESS}}\\
\textbf{Baseline} & 17.83 & 15.85 & 37.54 & 50.14 & 32.53 & 37.34 & 64.67 & 63.71 & 65.46 & 74.25 & 44.44 & 36.31 & 49.40 & 47.50 & 35.09 & 47.51 & 26.28 
% & 43.88 
\\
\hdashline
$(-)\ top- \tau$ & \cellcolor{blue!20}+17.56 & \cellcolor{blue!20}+18.48 & \cellcolor{blue!20}+6.37 & \cellcolor{blue!20}+0.09 & \cellcolor{red!20}-4.73 & \cellcolor{red!20}-4.51 & \cellcolor{red!20}-7.49 & \cellcolor{red!20}-8.04 & \cellcolor{red!20}-8.70 & \cellcolor{red!20}-11.94 & \cellcolor{blue!20}+3.44 & \cellcolor{blue!20}+10.91 & \cellcolor{blue!20}+1.50 & \cellcolor{blue!20}+0.28 & \cellcolor{blue!20}+1.49 & \cellcolor{blue!20}+4.57 & \cellcolor{red!20}-1.22 
% & \cellcolor{blue!20}+1.06 
\\
\hline
% \toprule
\multicolumn{18}{c}{\cellcolor[HTML]{DDDDDD}\textbf{HELPFUL}}\\
\textbf{Baseline} & 87.39 & 95.61 & 79.06 & 49.57 & 27.65 & 20.64 & 22.77 & 40.66 & 20.65 & 23.47 & 61.39 & 79.17 & 63.96 & 65.28 & 39.82 & 71.55 & 23.30 
% & 51.29 
\\
\hdashline
$(-)\ top-\tau$ & \cellcolor{red!20}-18.39 & \cellcolor{red!20}-21.58 & \cellcolor{red!20}-24.15 & \cellcolor{red!20}-4.57 & \cellcolor{red!20}-5.80 & \cellcolor{red!20}-1.49 & \cellcolor{blue!20}+11.90 & \cellcolor{blue!20}+5.56 & \cellcolor{blue!20}+10.75 & \cellcolor{blue!20}+7.67 & \cellcolor{red!20}-11.50 & \cellcolor{red!20}-24.21 & \cellcolor{red!20}-9.91 & \cellcolor{red!20}-14.17 & \cellcolor{red!20}-3.18 & \cellcolor{red!20}-21.66 & \cellcolor{blue!20}+0.66 
% & \cellcolor{red!20}-7.30 
\\
\hline
\multicolumn{18}{c}{\cellcolor[HTML]{DDDDDD}\textbf{HELPFUL-HARMLESS}} \\

\textbf{Baseline} & 88.44 & 94.15 & 76.67 & 48.66 & 27.18 & 18.84 & 22.34 & 43.38 & 20.89 & 21.35 & 57.17 & 72.02 & 64.11 & 59.03 & 40.12 & 78.93 & 22.64 
% & 50.35 
\\
\hdashline
$(-)\ top-\tau$ & \cellcolor{red!20}-24.89 & \cellcolor{red!20}-29.01 & \cellcolor{red!20}-23.04 & \cellcolor{red!20}-2.61 & \cellcolor{red!20}-2.11 & \cellcolor{blue!20}+1.32 & \cellcolor{blue!20}+14.92 & \cellcolor{blue!20}+4.61 & \cellcolor{blue!20}+17.03 & \cellcolor{blue!20}+17.16 & \cellcolor{red!20}-6.44 & \cellcolor{red!20}-23.81 & \cellcolor{red!20}-10.51 & \cellcolor{red!20}-8.33 & \cellcolor{red!20}-2.62 & \cellcolor{red!20}-24.68 & \cellcolor{blue!20}+3.12 
% & \cellcolor{red!20}-5.88 
\\
\toprule
% & \cellcolor{red!20}-6.70 
\end{tabular}
\end{adjustbox}
\caption{Average accuracy change when neurons are ablated, on Helpful-specific tasks relative to baseline ($\tau=0$) for $\tau=10$. Column abbreviations: 
Alp‑E/H/L--AlpacaEval (Easy/Hard/Length),
Fact--Factuality,
LB--LLM Bar (Inst--adversarial instruction, Out--adversarial output,
Man: manual adversary, Nbr--neighboring adversary, Nat--natural),
MT--MT-Bench (Easy/Hard/Medium),
Prec‑IF--Precise Instruction Following. (ref. Appendix~\ref{aba:abalation_behavious} for model-wise comparisons)}
\label{tab:ablation_p10_Helpful_main}
\end{table*}

\begin{table}[!th]
\centering
\setlength{\tabcolsep}{3pt}
\renewcommand{\arraystretch}{1.1}
\begin{adjustbox}{max width=\linewidth}
\begin{tabular}{lcccccccc}
\toprule
Model & \rotatebox{0}{DNA} &
\rotatebox{0}{RD} &
\rotatebox{0}{RO} &
\rotatebox{0}{Saf} &
\rotatebox{0}{SafR} &
\rotatebox{0}{SResp} &
\rotatebox{0}{XS-Ref} &
\rotatebox{0}{XS-Res}
% &Avg 
\\
\hline
\multicolumn{9}{c}{\cellcolor[HTML]{DDDDDD}\textbf{HARMLESS}}\\
\hline
\textbf{Baseline} & 88.73 & 95.78 & 99.00 & 69.68 & 97.39 & 11.56 & 98.67 & 5.76 
% & 70.82 
\\
\hdashline
$(-)\ top-\tau$ & \cellcolor{red!20}-22.10 & \cellcolor{red!20}-23.89 & \cellcolor{red!20}-27.39 & \cellcolor{red!20}-29.89 & \cellcolor{red!20}-23.97 & \cellcolor{blue!20}+24.57 & \cellcolor{red!20}-27.49 & \cellcolor{blue!20}+27.07 
% & 
% \cellcolor{red!20}-12.89 
\\
\hline
\multicolumn{9}{c}{\cellcolor[HTML]{DDDDDD}\textbf{HELPFUL}}\\
\hline
\textbf{Baseline} & 11.15 & 0.22 & 18.67 & 11.09 & 8.31 & 91.72 & 11.51 & 91.36 
% & 30.50 
\\
\hdashline
$(-)\ top-\tau$ & \cellcolor{blue!20}+16.26 & \cellcolor{blue!20}+13.83 & \cellcolor{blue!20}+16.39 & \cellcolor{blue!20}+7.49 & \cellcolor{blue!20}+14.70 & \cellcolor{red!20}-19.79 & \cellcolor{blue!20}+15.98 & \cellcolor{red!20}-21.24 
% & \cellcolor{blue!20}+5.45 
\\
\multicolumn{9}{c}{\cellcolor[HTML]{DDDDDD}\textbf{HELPFUL-HARMLESS}}\\
\textbf{Baseline} & 31.17 & 41.56 & 87.00 & 33.41 & 83.96 & 57.84 & 86.08 & 63.91 
% & 60.62 
\\
\hdashline
$(-)\ top-\tau$ & \cellcolor{blue!20}+5.84 & \cellcolor{red!20}-3.72 & \cellcolor{red!20}-33.39 & \cellcolor{red!20}-9.20 & \cellcolor{red!20}-29.02 & \cellcolor{blue!20}+0.63 & \cellcolor{red!20}-33.69 & \cellcolor{red!20}-2.87 \\
\toprule
\end{tabular}
\end{adjustbox}
\caption{
Average accuracy change when neurons are ablated, on harmless specific tasks relative to the baseline ($\tau=0$) for $\tau=10$.
Column abbreviations: DNA--Do Not Answer, 
RD--Refusals Dangerous, 
RO--Refusals Offensive, 
Saf--Safety, 
SafR--Safety Refuse, 
SResp--Safety Response, 
XS-Ref--XSTest Should Refuse, 
XS-Resp--XSTest Should Respond.
(ref. Appendix~\ref{aba:abalation_behavious} for model-wise comparisons)}
\label{tab:ablation_p10_Harmless_main}
\end{table}
\textbf{Neuron ablation.} To assess the importance of the identified objective-specific neurons, we ablate them in the corresponding reward models (RMs) and measure the resulting performance change.\footnote{We use the HH-RLHF test set. Each model is evaluated on the subset corresponding to its training data.} 
For neuron selection, we build on \citet{chen2025towards}, the closest state-of-the-art approach for identifying safety neurons in LLMs. We adapt it to the RM setting by selecting the top-$\tau$ neurons whose activation magnitudes differ most from those of the corresponding base RM. In addition, we include a random neuron selection baseline.

The selected neurons are ablated by setting their activations to zero during the forward pass~\cite{meng2022locating}. We also considered mean ablation, where neuron activations are replaced with their mean values. However, this intervention exhibited substantially higher variance across models, making analysis less reliable. We therefore adopt zero ablation in all experiments (see Appendix~\ref{aba:neural_abalation_comparison}).

The magnitude-based method consistently identifies the most influential neurons across all three RMs and for different values of $\tau$ (see Figure~\ref{fig:abalation_change}, App. for detailed results). 
% The stronger performance of zero over mean ablation and the improved performance of magnitude-based neuron selection compared to \citet{chen2025towards} highlight the need for the development of RM-specific interpretation techniques rather than transferring best practices of LLM interpretability. 
\textcolor{black}{We also analyse the distribution of objective-specific neurons in Appendix~\ref{aba:abalation_neuron_distribution}}.
% \begin{figure*}[!th]
% \centering
% \begin{minipage}{0.48\linewidth}
%     \centering
%     \includegraphics[width=\linewidth]{fig/abalation_effect.pdf}
%     \caption*{\textbf{(Left)} Effect of ablating neurons. Magnitude-based selection consistently yields the most informative neuron set.}
% \end{minipage}
% \hfill
% \begin{minipage}{0.48\linewidth}
%     \centering
%     \includegraphics[width=\linewidth]{fig/layerwise_distribution.pdf}
%     \caption*{\textbf{(Right)} Layer-wise distribution of selected neurons, showing a concentration in the upper layers of the LLM.}
% \end{minipage}
% \caption{\textbf{Effect and distribution of ablated neurons.} (Left) Performance impact of neuron ablation under different selection strategies. (Right) Layer-wise distribution of neurons selected for ablation, with most neurons residing in the top layers of the model.}
% \label{fig:ablation_and_distribution}
% \vspace{-5mm}
% \end{figure*}

\subsection{Characterising the Effect of Objective-specific Neurons}
\label{sec:4.2}

\textbf{Effect of ablation on helpfulness- and harmlessness-specific tasks.} We now examine how objective-specific neurons shape RM behaviour by ablating them from their corresponding RMs and evaluating the ablated RM's accuracy on both helpfulness‑ and harmlessness‑specific tasks. As shown in Table~\ref{tab:ablation_p10_Helpful_main} and Table~\ref{tab:ablation_p10_Harmless_main}, ablating harmlessness‑associated neurons from the harmless RM leads to an average accuracy drop of 12 points on harmlessness‑specific tasks, while slightly improving accuracy on helpfulness‑specific tasks. Similarly, ablating helpfulness‑specific neurons from the helpful RM decreases accuracy by 7 points on helpfulness‑specific tasks, while increasing accuracy by approximately 5 points on harmlessness‑specific tasks. The changes are also usually mirrored: if ablating harmlessness‑specific neurons reduces performance, ablating helpfulness‑specific neurons increases it, and vice versa. These results highlight the intrinsic tension between alignment objectives, showcasing that \textit{these neurons are not only essential for their target objective but also inhibit performance on the opposing alignment objective.}

Table~\ref{tab:ablation_p10_Harmless_main} and Table~\ref{tab:ablation_p10_Helpful_main} also show that, consistent with our earlier observations in \S\ref{sec:behavioural}, some tasks benefit from features learned for the opposing objective. In particular, for helpfulness‑specific tasks such as Safety (response) and XSTest (should respond), performance improves when harmlessness‑specific neurons are ablated and drops when helpfulness‑specific neurons are ablated. A similar pattern is observed for the LLM Bar variants in helpfulness-specific tasks.

Ablating mixed-objective neurons from the mixed-objective RM also decreases accuracy by $6\%$ on helpfulness-specific tasks. However, certain tasks (e.g. LLMBar variants, Focus, and Precise Instruction Following (IF)) are exceptions. They are specifically designed to penalise superficial helpfulness, which the single-objective helpful models also struggled to learn (\S\ref{sec:section1}).
Accuracy on harmlessness-specific tasks also declines by $13.12\%$ following ablation. An exception is the \textit{Do Not Answer} subset of RewardBench, which appears as an outlier and does not extend to other refusal-based tasks, where performance degrades.

\textbf{Score distribution analysis.} Figure~\ref{fig:scoring_abalated}, App., shows the distribution of reward scores before and after neuron ablation for both helpful and harmless RMs. Ablating objective‑specific neurons shifts score distributions, with scores for preferred and rejected responses becoming weaker and more concentrated around zero, indicating reduced model confidence across all outputs. The compressed score range also suggests that the ablated neurons contribute substantially to the reward signal's expressiveness. \textit{Overall, neuron ablation affects model behaviour by changing the strength and separability of reward scores, further underscoring the importance of these neurons.}\looseness=-1

\section{Mixed Objective Conflict}
\label{sec:section3}
\subsection{Do the objectives use similar neurons?}
\label{sec:5.1}

\textcolor{black}{Given the importance of objective-specific neurons in shaping the behaviour and scoring of RMs, we extend the notion of \textit{retention} to measure the extent to which objective-specific neurons are shared between two reward models. Formally, let $\omega_1$ and $\omega_2$ denote the sets of objective-specific neurons identified in $RM_1$ and $RM_2$, respectively. We define retention as the intersection-over-union percentage of these two sets, representing the proportion of objective-specific neurons shared by both RMs.}

\textcolor{black}{We find the retention between the harmless-RM and mixed-RMs to be $50\%$ and that of helpful-RM and mixed-RM to be $60\%$; this indicates a substantial overlap between single-objective and mixed-RMs.} This aligns with the intuition that the concepts used by the single- and mixed- RMs should overlap to some extent, while additional neurons also become important to accommodate the joint optimisation of the mixed objective. 

However, we surprisingly also observe that the retention between the single-objective RM neurons themselves is approximately $50\%$. These results suggest that, despite behavioural differences, the models rely on a shared subset of underlying concepts encoded by the shared neurons. \textcolor{black}{To validate that large overlap is not due to chance, we constructed a layer-preserving null hypothesis. For each layer, we randomly sampled the same proportion of neurons as identified in the single-objective models and computed the resulting retention across five random seeds. This null hypothesis yielded an average retention of only 30\%, substantially below the observed 50\% retention, suggesting that the retention reflects genuine sharing of neurons across objectives rather than due to chance. This is surprising, as the two objectives exhibit distinct behavioural characteristics (see \S\ref{sec:section1}).} 

\subsection{Do the shared neurons drive mixed RM behaviour?}
\label{sec:5.2}
\textcolor{black}{To fully investigate the role of the shared neurons, we partition the helpfulness- and harmlessness-specific neuron sets into three groups: \emph{helpfulness-only} (i.e., $\omega_{H}-\omega_{S}$), \emph{harmlessness-only} (i.e., $\omega_{S}-\omega_{H}$), and \emph{shared} neurons (i.e., $\omega_{H}\cap\omega_{S}$). Henceforth, we refer to the two sets $(\omega_{H}-\omega_{S})$ and $(\omega_{S}-\omega_{H})$ as unique neuron sets. We then perform targeted ablations in the mixed-objective RM to assess the functional importance of the three neuron groups.}

\begin{table}[!th]
\centering
\small
\begin{tabular}{l|cc}
\toprule
\textbf{Setting} & \textbf{Harmless Tasks} & \textbf{Helpful Tasks} \\
\multicolumn{3}{c}{\cellcolor[HTML]{DDDDDD}\textbf{Accuracy}}\\
\hline
\textbf{Baseline}        & 61.04 & 50.47 \\
\hdashline
$(-)\ shared$            &  \cellcolor{red!20}-10.00 &  \cellcolor{red!20}-4.40 \\
$(-)\ harmless\ only$    &  \cellcolor{red!20}-1.18  &  \cellcolor{red!20}-0.76 \\
$(-)\ helpful\ only$     &  \cellcolor{red!20}-0.00  &  \cellcolor{red!20}-0.83 \\
% \hline
% \multicolumn{3}{c}{\cellcolor[HTML]{DDDDDD}\textbf{R}}\\
% \hline
% \textbf{Baseline}        & 0.6285 & 0.5262 \\
% \hdashline
% $(-)\ shared$            &  \cellcolor{red!20}-0.0306 &  \cellcolor{red!20}-0.0127 \\
% $(-)\ harmless\ only$    &  \cellcolor{red!20}-0.0043  &  \cellcolor{red!20}-0.0304 \\
% $(-)\ helpful\ only$     &  \cellcolor{red!20}-0.0001  &  \cellcolor{red!20}- 0.0321\\
\toprule
\end{tabular}
\caption{Average accuracy change relative to baseline ($\tau=0$) for mixed-objective model, separated by task orientation. (ref. App.~\ref{aba:shared_neurons} for model-wise details.)}
\label{tab:ablation_avg}
% \vspace{-5mm}
\end{table}

Table~\ref{tab:ablation_avg} shows that \textit{shared neurons are critical in determining RM behaviour}. In particular, ablating shared neurons results in accuracy degradation, with an average drop of 10\% points on harmlessness-specific tasks and 4.4\% points on helpfulness-specific tasks. Notably, this degradation is comparable to that observed when ablating the top $10$ neurons identified in earlier sections (\S\ref{sec:4.2}).
% , despite shared neurons constituting only an average of $4.75\%$ of the total neuron set.
Furthermore, this effect is substantially larger than that observed when ablating objective-only neurons.

\subsection{Do shared neurons encode conflicting objective-specific representations?} 

\textcolor{black}{We showed that a substantial fraction of objective-specific neurons are shared across objectives and that these neurons play a disproportionately important role in determining the behaviour of the mixed-objective RM. However, their functional role remains unclear. Do these neurons encode objective-agnostic representations that are useful for both objectives, or do they contain objective-specific information learned by both models?} 

\textcolor{black}{To answer this question, we perform an activation-patching analysis. We consider two competing hypotheses. First, shared neurons may encode common semantic features that both objectives rely on. In this case, patching activations from either single-objective RM into the mixed-objective RM should produce behavioural changes in similar directions. Alternatively, the two objectives may have learned incompatible objective-specific representations within the same neurons. Under this hypothesis, patching activations from the helpfulness and harmlessness RMs should induce conflicting behavioural effects. In other words, we test whether shared neurons cooperate by encoding common features or compete by encoding objective-specific features that push behaviour in opposing directions.}

\textcolor{black}{We consider two patching strategies. In the \emph{shared method}, activations are patched only within the shared neuron set. In the \emph{unique method}, activations are patched within the unique neuron sets. For each method, we separately patch activations from the single-objective RM into the mixed-objective RM. We measure the directional agreement of patching from the two single-objective RMs. We assign a score of $+1$ when both interventions produce changes in the same direction, $-1$ when they produce changes in opposite directions, and $0$ when neither intervention has an effect. As a baseline, we repeat the analysis using randomly sampled neurons.}

\begin{table}[!th] 
\centering \small 
\begin{adjustbox}{max width=\linewidth} 
\begin{tabular}{l|cc} 
\toprule \textbf{Setting} & \textbf{Harmless Tasks} & \textbf{Helpful Tasks} \\ 

\midrule \multicolumn{3}{c}{\cellcolor[HTML]{DDDDDD}\textbf{Directional Change Score}} \\ \hline 
Shared Method (Rand) & 0.375 & 0.088 \\ \hdashline 
Shared Method & -0.500 & -0.039 \\ \hline 
Unique Method (Rand) & 0.000 & 0.019 \\ \hdashline 
Unique Method & 0.375 & 0.029 \\ \bottomrule 
\end{tabular} 
\end{adjustbox} 
\caption{Average directional agreement score by task category (see App.~\ref{aba:task_wise_direction_chage} for more details. Random abbreviated to Rand.)} 
\label{tab:patching_avg} \end{table}

\textcolor{black}{As shown in Table~\ref{tab:patching_avg}, the unique method yields positive directional agreement scores that resemble the random-patching baseline, indicating that patching of unique neurons affects behaviour in similar ways. In contrast, the shared method produces substantially lower agreement and negative scores, implying that the same shared neurons can drive the mixed-objective RM in opposing directions depending on the single-objective RM. These results suggest that shared neurons encode some objective-specific representations inherited from both helpfulness and harmlessness training. Combined with the ablation results, this provides evidence that interference within the shared neuron subspace is a key factor underlying the performance degradation observed in mixed-objective RMs. It also shows that shared neurons are areas of representational conflict, rather than reused representations.}
\looseness=-1

\section{Related Work}

\textbf{RM Analysis.}
RMs are a central component of RLHF pipelines, where they are trained to approximate human preferences over model outputs~\cite{ouyang2022traininglanguagemodelsfollow,bai2022training}. Prior work has primarily focused on improving RM robustness, mitigating reward hacking, and designing better training objectives for alignment~\cite{dong2023steerlm,10.5555/3666122.3666604,wang-etal-2024-interpretable}. Recent benchmarks such as RewardBench~\cite{lambert2024rewardbench}, RewardBench2 ~\cite{malik2025rewardbench2advancingreward}, and RM-Bench~\cite{liu2025rmbench} further study RM generalisation and calibration across diverse evaluation settings. While these studies highlight challenges such as misalignment and generalisation gaps, they largely treat RMs as black boxes, leaving their internal mechanisms under-explored.

\textbf{Alignment Trade-offs and Multi-Objective Learning.}
The tension between helpfulness and harmlessness has been widely recognised as a key challenge in alignment. Mitigations include balancing objectives through data curation, modified loss functions, post-training calibration, or Pareto-optimal RM weight interpolation~\cite{bai2022constitutionalaiharmlessnessai,sun2024salmon,rame2023rewarded}. 
% Related work in multi-objective optimization also studies trade-offs between competing objectives in machine learning systems, often focusing on Pareto optimality or scalarization strategies. 
However, these approaches primarily characterise the tension behaviourally through outputs, refusal rates, or optimisation trade-offs, without examining how competing objectives are represented internally in RMs.\looseness=-1
% While existing work has proposed to mitigate alignment tension in LLMs by modifying there training regimes~\citep{dong2023steerlm,10.5555/3666122.3666604,wang2024helpsteer,hong2025on,wang2024secrets,sun2024salmon}, l

\textbf{Mechanistic Interpretability.}
Recent advances in mechanistic interpretability aim to uncover how transformer models represent and process information by analysing neurons, attention heads, and circuits~\cite{NEURIPS2020_92650b2e,meng2022locating,geva-etal-2021-transformer}. 
Prior work has localised reasoning behaviours and safety-relevant representations within language models, often using activation analysis and causal interventions such as ablations~\cite{zhou2024role,chen2025towards}. More recent work has begun exploring interpretability for RMs, including contrastive explanations~\cite{jiang2025interpreting} and sparse-autoencoder approaches that attribute reward scores to interpretable feature activations~\cite{zhang2026interpretable,li2025safer}. However, existing work does not study how competing alignment objectives, such as helpfulness and harmlessness are jointly represented within RMs, nor how shared representations contribute to alignment interference during mixed-objective training.\looseness=-1

% Notes
% Single objective model -> mixed objective paching -> A B shared
% Sample patching -> Mixed objective patching -> C D random
% Sample patching -> Mixed objective patching -> E F not shared
% Number of taks whoes perforamnce changes in opposite diffrection is significant w.r.t. random patching
%

% Activation-based methods have been used to identify neurons associated with specific concepts, while intervention techniques such as ablation and causal tracing have provided insights into model behaviour. Although these methods have been successfully applied to language modelling, their application to reward models and alignment remains limited.

\section{Conclusion}

In this work, we investigated the interaction between helpfulness and harmlessness within RMs through a behavioural analysis and interpretability lens. Our results show that mixed-objective training introduces alignment tension, leading to degraded performance and limited retention of capabilities learned by single-objective RMs. \textcolor{black}{Through neuron analysis, we found objective-specific neurons that are responsible for the behaviour of the RMs. Further analysis of these neurons shows that
% targeted ablations, and activation patching, we find 
that neurons associated with the two objectives are partially shared. The patching and ablation results of these shared neurones showed that shared representations play a central role in encoding the behaviour of the RM and in encoding objective-specific information. We believe our findings fill a much needed gap in reward model interpretive research and can be use to build better reward models. One possible avenue to do so is by disentangling the objective-specific information encoded in the shared neurons to help reduce alignment tension in the mixed model.}\looseness=-1

\section{Limitations}

Our analysis focused on understanding the reward models used in RLHF by training single- and mixed-objective-specific models. This approach helped us decouple the tension between alignment objectives and analyse the behavioural changes each objective introduces in the reward model, showcasing the feasibility of such analysis. We believe future work can expand on this type of analysis to study more recent RLHF training techniques such as Direct Preference Optimisation (DPO)~\cite{rafailov2023direct} and RLOO~\cite{ahmadian-etal-2024-back}. Furthermore, while we focused on broader definitions of helpfulness and harmlessness, recent datasets have introduced more fine-grained distinctions that could provide deeper insights into this analysis~\cite{beavertails}.

Additionally, our experimental setup was limited by computational constraints, as we chose not to use parameter-efficient fine-tuning techniques in order to preserve interpretability. Future work could investigate how such techniques affect model structure and extend the analysis to larger-scale models and other model families.

Future work should also focus on understanding the representation of shared neurons, as they may exhibit interesting polysemantic behaviors~\cite{gurnee2023finding}. This could be further explored by modifying the ablation pipeline to specifically target such neurons. We believe that understanding and disentangling these neurons could contribute to the development of more principled and better-designed reward models.

% relies on activation-based neuron importance and controlled ablations, which provide a useful but approximate view of internal mechanisms. Additionally, experiments are conducted on a limited set of datasets and model families, and further validation across broader settings would strengthen the generality of our findings.

\bibliography{custom}

\newpage
\appendix

\section{Appendix}
\label{sec:appendix}

\subsection{Overview of LLM architecture}
\label{aba:overview_LLMs}
\noindent \textbf{Large Language Models:} Autoregressive transformer layers~\cite{NIPS2017_3f5ee243} form the backbone of LLMs. Within each transformer block, there is a multi-head self-attention (MHA) module, which facilitates the movement of information across token positions~\cite{wang2023interpretability}, followed by a multi-layer perceptron (MLP) that applies non-linear transformations to token representations and facilitates memorisation of concepts in the LLMs~\cite{geva-etal-2021-transformer}.

For an input sequence $\mathbf{w}=\langle w_0,\ldots,w_T\rangle$, each token $w_i$ is first mapped by the model into an embedding $\mathbf{h}_i\in\mathbb{R}^d$ using an embedding matrix $W_E$. This sequence of hidden states $\{\mathbf{h}^{l}\}$ is then refined by each transformer layer at level $l$ as follows:
\[
\mathbf{h}^{l+1} = \mathbf{h}^{l}
+ \mathrm{MHA}^{l}(\mathbf{h}^{l})
+ \mathrm{MLP}^{l}\big(\mathbf{h}^{l} + \mathrm{MHA}^{l}(\mathbf{h}^{l})\big).
\]

Where the MLP can be seen as a gated feed-forward network, which can be written as:
\[
\mathrm{MLP}(\mathbf{h}^{l})
= \mathbf{W}_{\text{down}}^{\top}
\big(\sigma(\mathbf{W}_{\text{gate}}\mathbf{h}_i^{l}) \odot \mathbf{W}_{\text{up}}\mathbf{h}_i^{l}\big),
\]
where $\mathbf{W}_{\text{up}} \in \mathbb{R}^{d_m \times d}$ and $\mathbf{W}_{\text{down}}$ act as the key--value pair that store information, while the activation, $\mathbf{A} = \sigma(\mathbf{W}_{\text{gate}}\mathbf{h}^{l}) \in \mathbb{R}^{d_m}$, controls this interaction~\cite{gurnee2023finding}. Consequently, we can view an MLP in layer $l$ as comprising $d_m$ neurons, where the $j$-th neuron is activated if  $a_j > 0$.
\subsection{Model Card.}
\label{aba:model_card}

\begin{table}[h]
\centering
\small
\begin{adjustbox}{max width=\linewidth}
\begin{tabular}{l l}
\toprule
\textbf{Parameter} & \textbf{Value} \\
\hline
learning\_rate & $5 \times 10^{-5}$ \\
per\_device\_train\_batch\_size & $8$ \\
per\_device\_eval\_batch\_size & $8$ \\
lr\_scheduler\_type & linear \\
dataset size & 161k\\
\toprule
\end{tabular}
\end{adjustbox}
\caption{Hyper-parameters for RM training.}
\label{tab:traning_values}
\end{table}

\begin{table}[h]
\centering
\small
\begin{adjustbox}{max width=\linewidth}
\begin{tabular}{l l}
\toprule
\textbf{Model} & \textbf{Hugging Face Model} \\
\hline
SmoLLM2-1.7b & HuggingFaceTB/SmoLLM2-1.7B \\
SmoLLM2-135m & HuggingFaceTB/SmoLLM2-135M \\
SmoLLM2-360m & HuggingFaceTB/SmoLLM2-360M \\
GPT2-large & openai-community/GPT2-large \\
GPT2-medium & openai-community/GPT2-medium \\
GPT2-small & openai-community/GPT2 \\
\toprule
\end{tabular}
\end{adjustbox}
\caption{Hugging Face repositories for all models used in our experiments.}
\label{tab:hf_models}
\end{table}

\begin{table}[h]
\centering
\small
\begin{tabular}{lc}
\toprule
\multicolumn{2}{c}{\textbf{Helpfulness-only}} \\
\hline
GPT2-large    & $69.07_{1.12}$ \\
GPT2-medium   & $71.31_{0.65}$ \\
GPT2-small    & $69.20_{0.68}$ \\
SmoLLM2-1.7b  & $74.53_{0.21}$ \\
SmoLLM2-135m  & $70.32_{0.18}$ \\
SmoLLM2-360m  & $72.76_{0.30}$ \\

\hline
\multicolumn{2}{c}{\textbf{Harmless-only}} \\
\hline
GPT2-large    & $72.10_{0.66}$ \\
GPT2-medium   & $72.04_{0.08}$ \\
GPT2-small    & $70.26_{0.32}$ \\
SmoLLM2-1.7b  & $73.75_{0.30}$ \\
SmoLLM2-135m  & $70.62_{0.62}$ \\
SmoLLM2-360m  & $72.28_{0.31}$ \\
\hline
\multicolumn{2}{c}{\textbf{Harmless and harmless}} \\
\hline
GPT2-large    & $64.67_{3.5}$ \\
GPT2-medium   & $68.17_{0.12}$ \\
GPT2-small    & $64.89_{0.41}$ \\
SmoLLM2-1.7b  & $73.17_{0.26}$ \\
SmoLLM2-135m  & $64.42_{0.62}$ \\
SmoLLM2-360m  & $68.54_{0.23}$ \\
\toprule
\end{tabular}
\caption{\textcolor{black}{Accuracy of the model on the validation set components corresponding to the HH-RLHF subsets it was trained on.}}
\label{tab:traning_hhrlhf}
\end{table}

Table~\ref{tab:traning_values} and Table~\ref{tab:hf_models} provide details about the hyper-parameters and Hugging Face repositories used for training. \textcolor{black}{We trained our RM for two epochs and selected the best-performing RM based on its validation accuracy, which was evaluated every 500 training steps (Table~\ref{tab:traning_hhrlhf})}. Furthermore,  we avoid parameter‑efficient fine‑tuning methods to preserve interpretability, and instead use full‑parameter fine‑tuning. While this limits the scale of reward models we can train, we consider it a necessary trade‑off for this first‑principled analysis of internal representations of reward models.
\subsection{Evaluation Dataset.}
\label{aba:eval_dataset}

Table~\ref{tab:rewardbench_datasets},~\ref{tab:rewardbench_datasets2}, and~\ref{tab:rm-bench} contain details about RewardBench, RewardBench 2 and RM-Bench datasets.
\begin{table*}[h]
\centering
\small
\begin{tabularx}{\textwidth}{l c X}
\toprule
\textbf{Subset} & \textbf{N} & \textbf{Description} \\
\hline

AlpacaEval Easy & 100 & To analyse instruction-following ability using outputs from GPT-4 Turbo and Alpaca 7B~\cite{alpaca_eval} \\
AlpacaEval Length & 95 & Responses extracted from LLaMA-2 Chat 70B and Guanaco 13B completions \\
AlpacaEval Hard & 95 & Responses extracted from Tulu-2 DPO 70B vs. Davinci-003 completions \\
MT Bench Easy & 28 & Responses extracted from MT Bench that are scored 10 and 1~\cite{10.5555/3666122.3668142} \\
MT Bench Medium & 40 & Responses extracted from MT Bench completions rated 9 vs. 2--5 \\
MT Bench Hard & 37 & Responses extracted from MT Bench completions rated 7--8 vs. 5--6 \\
LLMBar Natural & 100 & Comparisons from~\citet{zeng2024evaluating} \\
LLMBar Adversarial Neighbor & 134 & Tougher comparisons extracted via semantically similar prompts \\
LLMBar Adversarial GPTInst & 92 & GPT-4 generated similar prompt comparisons \\
LLMBar Adversarial GPTOut & 47 & GPT-4 generated intentionally unhelpful responses \\
LLMBar Adversarial Manual & 46 & Manually curated adversarial examples \\

Refusals Dangerous & 100 & Dangerous queries that should be refused \\
Refusals Offensive & 100 & Offensive queries that should be refused \\
XSTest Should Refuse & 154 & Prompts that should be refused~\cite{rottger-etal-2024-xstest} \\
XSTest Should Respond & 250 & Responses that should be preferred despite containing trigger words~\cite{rottger-etal-2024-xstest} \\
Do Not Answer & 136 & Contains queries that models should refuse~\citet{wang-etal-2024-answer} \\
PRM Math & 447 & Comparisons between human and incorrect LLM math answers~\cite{lightman2024let} \\

\toprule
\end{tabularx}
\caption{Summary of Reward Bench. (Adversarial abbreviated as Adver).}
\label{tab:rewardbench_datasets}
\end{table*}

\begin{table*}[h]
\centering
\small
\begin{tabularx}{\textwidth}{l c X}
\toprule
\textbf{Subset} & \textbf{N} & \textbf{Description} \\
\hline
Factuality&475& Tests the ability of RM to detect hallucinations and other errors in generations.\\
Precise IF &160& Evaluated the ability of RM select generations that follow precise instructions of prompt.\\
Math & 183 & Evaluates RM's mathematical abilities.\\
Safety&450& Tests RM's ability to comply or refuse prompts that are harmful to use cases.\\
Focus&495& Evaluated RM ability to select highly specific on topic answers.\\
\hline
\end{tabularx}
\caption{Summary of RewardBench2.}
\label{tab:rewardbench_datasets2}
\end{table*}

\begin{table*}[h]
\centering
\small
\begin{tabularx}{\textwidth}{l c X}
\hline
\textbf{Subset} & \textbf{N} & \textbf{Description} \\
\hline
Chat&519& Tests the ability of RM to detect hallucinations and other errors in generations.\\
Safety (should Response)&157& Alarming but benign prompts collected from~\citet{rottger-etal-2024-xstest} that should be responded.\\
Safety (should Refuse)& 284 &Harmful prompts collected from~\citet{rottger-etal-2024-xstest} that should be refused.\\
\toprule
\end{tabularx}
\caption{Summary of RM-Bench.}
\label{tab:rm-bench}
\end{table*}

\begin{figure}[!t]
\begin{center}
\includegraphics[width=\linewidth]{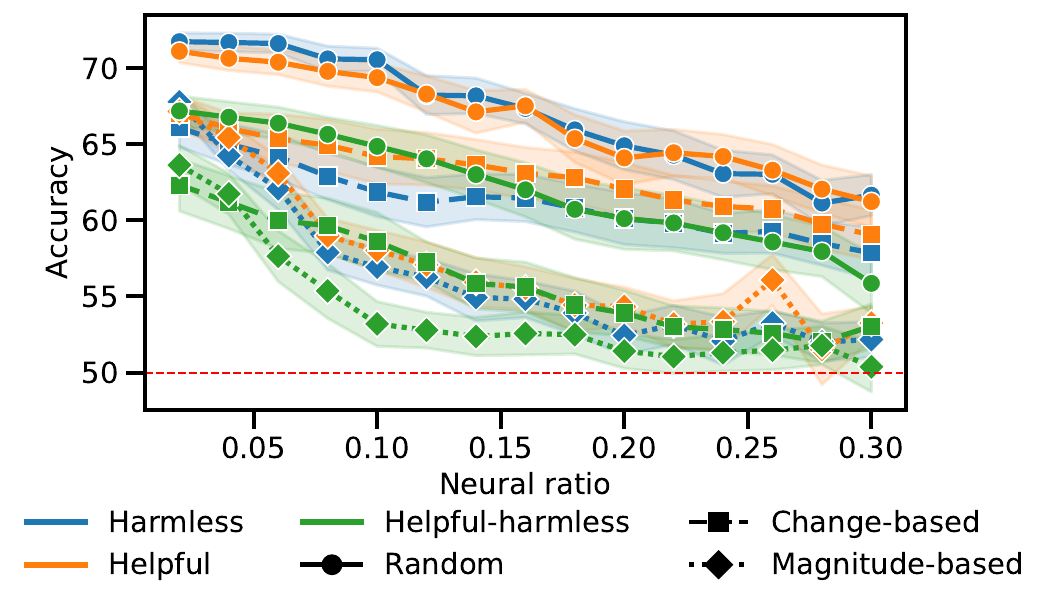}
\caption{\textbf{Effect of ablating neurons.}
Magnitude-based selection consistently yields the best set of neurons. The change-based method follows \citet{chen2025towards}, while the random baseline samples neurons uniformly. Our approach is magnitude-based (see Appendix~\ref{aba:model_wise_abalation_effect} for model-wise trends).}
% \caption{\textbf{Effect of ablating neurons.} Magnitude-based selection consistently yields the best set of neurons. The change-based method refers to the selection strategy proposed by \citet{chen2025towards}, while the random baseline selects neurons uniformly at random, and our approach is refeed as magnitude-based. (ref. Appendix~\ref{aba:model_wise_abalation_effect} for model-wise trends). }
\label{fig:abalation_change}
\end{center}
\vspace{-5mm}
\end{figure} 
\subsection{Model-wise effect of ablation on performance.}
\label{aba:model_wise_abalation_effect}

\begin{figure*}[!th]
\begin{center}
\includegraphics[width=\textwidth]{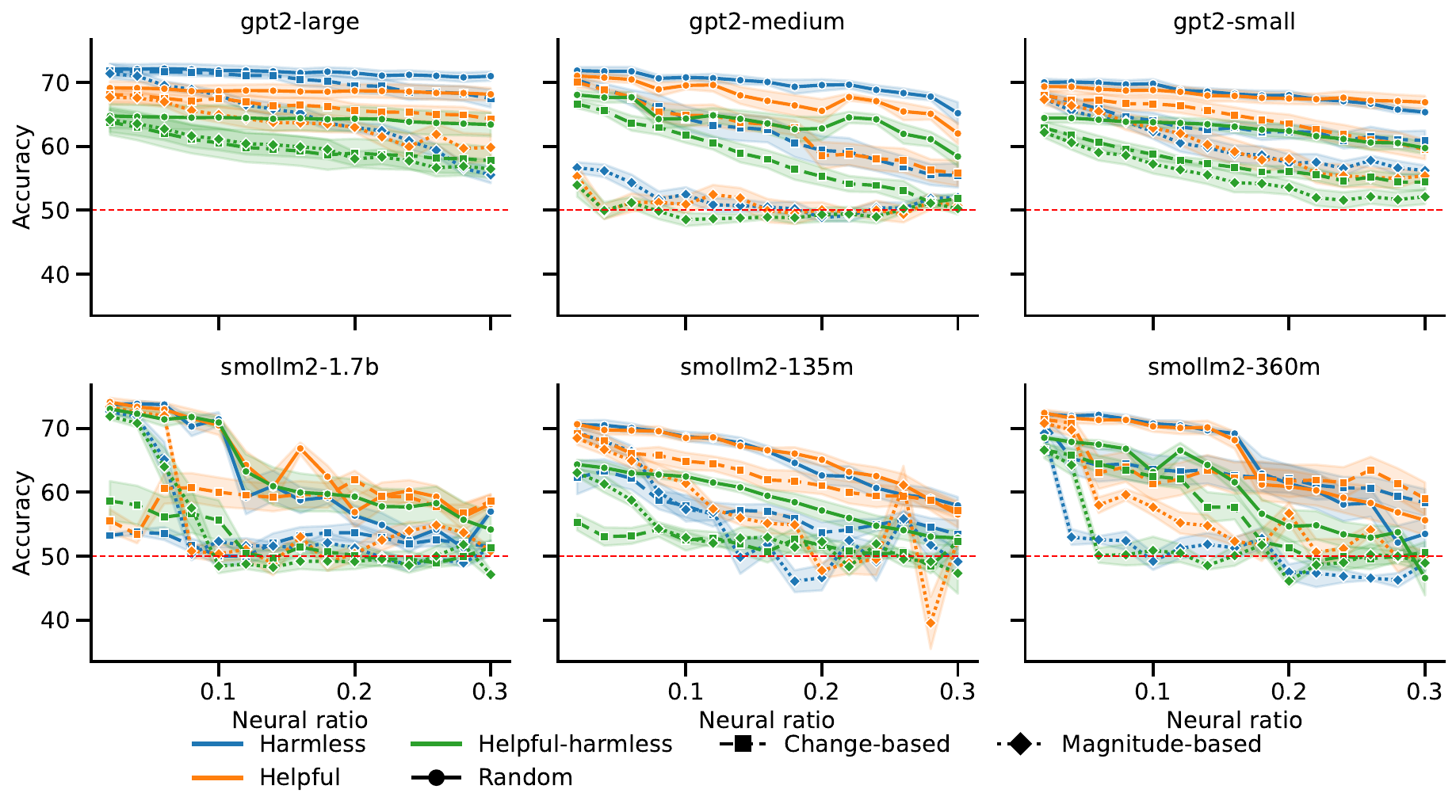}
\caption{\textbf{Effect of ablating neurons on each model.} Magnitude based
selection consistently gives the best set of neurons.}
\label{fig:gride_aba}
\end{center}
\vspace{-5mm}
\end{figure*}

Figure~\ref{fig:gride_aba} illustrates the effect of neuron ablation across different models. Our approach consistently identifies neurons of interest in all models, with the exception of one. In the case of GPT‑2 Large, the impact of ablation is minimal; however, this observation aligns with the results obtained from change-based neuron scoring. We therefore attribute this behaviour to a model-specific characteristic rather than a limitation of our method.

\subsection{Pairwise retention score}
\label{aba:pair_wise_retention}

Table~\ref{tab:HH-harmless_retrention} shows the retention score between the Helpful-harmless and harmless RM. The model seems to retrain refusal-oriented tasks more than others. Similarly, table~\ref{tab:HH-helpful_retrention} shows the score between the Helpful-harmless and helpful RM. Here we find the model to have high values of retention for Alpaca tasks.

\begin{table}[t]
\centering
\begin{tabular}{l c}
\hline
\textbf{Sub-task} & \textbf{HH-HA retention} \\
\hline
Alpaca Eval (Easy) & $0.052_{0.005}$ \\
Alpaca Eval (Hard) & $0.102_{0.014}$ \\
Alpaca Eval (Length) & $0.205_{0.064}$ \\
Chat & $0.213_{0.016}$ \\
Do Not Answer & $0.280_{0.034}$ \\
Factuality & $0.087_{0.016}$ \\
Focus & $0.045_{0.026}$ \\
LLM Bar (Adver GPTInst) & $0.122_{0.016}$ \\
LLM Bar (Adver GPTOut) & $0.168_{0.020}$ \\
LLM Bar (Adver Manual) & $0.070_{0.049}$ \\
LLM Bar (Adver Neighbor) & $0.156_{0.017}$ \\
LLM Bar (Natural) & $0.227_{0.048}$ \\
MT Bench (Easy) & $0.152_{0.090}$ \\
MT Bench (Hard) & $0.213_{0.048}$ \\
MT Bench (Med) & $0.225_{0.079}$ \\
Math & $0.236_{0.021}$ \\
Math PRM & $0.679_{0.046}$ \\
Precise IF & $0.112_{0.010}$ \\
Refusals Dangerous & $0.443_{0.042}$ \\
Refusals Offensive & $0.840_{0.094}$ \\
Safety & $0.313_{0.044}$ \\
Safety Refuse & $0.788_{0.056}$ \\
Safety Response & $0.064_{0.017}$ \\
XStest Should Refuse & $0.874_{0.029}$ \\
XStest Should Respond & $0.024_{0.008}$ \\
\hline
\end{tabular}
\caption{Helpful-harmless and Harmless RM retention score. Variance is shown as a subscript.}
\label{tab:HH-harmless_retrention}
\end{table}

\begin{table}[t]
\centering
\begin{tabular}{l c}
\hline
\textbf{Sub-task} & \textbf{HH-HE retention} \\
\hline
Alpaca Eval (Easy) & $0.838_{0.032}$ \\
Alpaca Eval (Hard) & $0.898_{0.037}$ \\
Alpaca Eval (Length) & $0.731_{0.095}$ \\
Chat & $0.547_{0.068}$ \\
Do Not Answer & $0.233_{0.070}$ \\
Factuality & $0.407_{0.065}$ \\
Focus & $0.260_{0.035}$ \\
LLM Bar (Adver GPTInst) & $0.518_{0.098}$ \\
LLM Bar (Adver GPTOut) & $0.542_{0.170}$ \\
LLM Bar (Adver Manual) & $0.521_{0.184}$ \\
LLM Bar (Adver Neighbor) & $0.384_{0.056}$ \\
LLM Bar (Natural) & $0.586_{0.057}$ \\
MT Bench (Easy) & $0.661_{0.140}$ \\
MT Bench (Hard) & $0.734_{0.091}$ \\
MT Bench (Med) & $0.571_{0.085}$ \\
Math & $0.494_{0.061}$ \\
Math PRM & $0.761_{0.065}$ \\
Precise IF & $0.453_{0.094}$ \\
Refusals Dangerous & $0.000_{0.000}$ \\
Refusals Offensive & $0.160_{0.049}$ \\
Safety & $0.150_{0.074}$ \\
Safety Refuse & $0.053_{0.020}$ \\
Safety Response & $0.584_{0.065}$ \\
XStest Should Refuse & $0.117_{0.069}$ \\
XStest Should Respond & $0.637_{0.027}$ \\
\hline
\end{tabular}
\caption{Helpful-harmless and Helpful retention score. Variance shown as subscript.}
\label{tab:HH-helpful_retrention}
\end{table}

\subsection{Model-wise analysis of alignment tension.}
\label{aba:model_wise_alignment_tension}

Table~\ref{tab:behavioural_analysis_harmless} and~\ref{tab:behavioural_analysis_helpful} show the model wise performance on helpful and harmless tasks

\begin{table*}[!th]
\centering
\setlength{\tabcolsep}{3pt}
\renewcommand{\arraystretch}{1.1}
\begin{adjustbox}{max width=\textwidth}
\begin{tabular}{l|ccccccccccccccccc}
\toprule
Model &
\rotatebox{0}{Alp-E} &
\rotatebox{0}{Alp-H} &
\rotatebox{0}{Alp-L} &
\rotatebox{0}{Chat} &
\rotatebox{0}{Fact} &
\rotatebox{0}{Focus} &
\rotatebox{0}{LB-Inst} &
\rotatebox{0}{LB-Out} &
\rotatebox{0}{LB-Man} &
\rotatebox{0}{LB-Nbr} &
\rotatebox{0}{LB-Nat} &
\rotatebox{0}{MT-E} &
\rotatebox{0}{MT-H} &
\rotatebox{0}{MT-M} &
\rotatebox{0}{Math} &
\rotatebox{0}{Math-PRM} &
\rotatebox{0}{Prec-IF}
% & Avg 
\\
\hline
\multicolumn{18}{c}{\cellcolor[HTML]{DDDDDD}\textbf{HARMLESS}}\\

GPT2-large & 8.00$_{1.41}$ & 10.53$_{1.49}$ & 34.74$_{4.79}$ & 53.49$_{1.38}$ & 30.81$_{0.26}$ & 45.66$_{1.03}$ & 76.45$_{2.23}$ & 65.96$_{3.01}$ & 67.39$_{3.07}$ & 75.37$_{1.61}$ & 45.00$_{3.74}$ & 41.67$_{4.45}$ & 52.25$_{2.55}$ & 53.33$_{2.36}$ & 35.88$_{1.51}$ & 74.27$_{7.20}$ & 25.21$_{1.64}$ \\
GPT2-medium & 6.67$_{1.25}$ & 8.77$_{2.76}$ & 40.35$_{3.02}$ & 53.49$_{1.52}$ & 30.67$_{0.26}$ & 46.33$_{0.50}$ & 74.28$_{2.23}$ & 60.28$_{1.00}$ & 68.84$_{4.47}$ & 76.87$_{2.20}$ & 46.33$_{0.47}$ & 36.90$_{6.73}$ & 50.45$_{1.27}$ & 49.17$_{3.12}$ & 35.34$_{1.67}$ & 76.36$_{9.13}$ & 22.50$_{1.84}$ \\
GPT2-small & 7.33$_{2.62}$ & 7.72$_{2.16}$ & 39.30$_{1.31}$ & 52.02$_{2.11}$ & 32.14$_{1.34}$ & 45.39$_{0.25}$ & 69.57$_{2.35}$ & 65.96$_{4.60}$ & 68.12$_{2.71}$ & 78.86$_{0.93}$ & 41.33$_{1.89}$ & 46.43$_{7.72}$ & 56.76$_{2.21}$ & 57.50$_{2.04}$ & 35.37$_{1.88}$ & 64.28$_{2.59}$ & 23.75$_{0.51}$ \\
SmoLLM2-1.7b & 21.67$_{6.34}$ & 17.89$_{3.94}$ & 37.19$_{2.76}$ & 48.32$_{0.76}$ & 33.82$_{3.62}$ & 30.91$_{2.31}$ & 54.35$_{7.74}$ & 65.25$_{4.01}$ & 58.70$_{6.40}$ & 72.89$_{3.57}$ & 43.67$_{3.40}$ & 38.10$_{6.07}$ & 37.84$_{3.82}$ & 41.67$_{2.36}$ & 34.58$_{0.97}$ & 32.44$_{6.65}$ & 30.83$_{1.64}$ \\
SmoLLM2-360m & 34.67$_{14.38}$ & 25.26$_{7.64}$ & 36.49$_{6.51}$ & 46.08$_{2.81}$ & 32.49$_{1.90}$ & 26.60$_{6.20}$ & 55.80$_{10.70}$ & 63.12$_{5.01}$ & 68.12$_{2.71}$ & 71.39$_{5.53}$ & 47.33$_{0.94}$ & 29.76$_{6.07}$ & 45.05$_{4.59}$ & 38.33$_{5.14}$ & 33.57$_{1.83}$ & 18.94$_{10.46}$ & 27.92$_{1.56}$ \\
\hline
\multicolumn{18}{c}{\cellcolor[HTML]{DDDDDD}\textbf{HELPFUL}}\\
GPT2-large & 81.00$_{6.16}$ & 91.23$_{3.25}$ & 73.33$_{6.68}$ & 45.48$_{3.19}$ & 23.23$_{3.12}$ & 16.50$_{5.53}$ & 28.26$_{2.66}$ & 43.26$_{2.01}$ & 18.12$_{2.71}$ & 20.15$_{5.31}$ & 53.67$_{1.89}$ & 67.86$_{11.66}$ & 59.46$_{5.84}$ & 53.33$_{8.25}$ & 38.99$_{1.86}$ & 77.85$_{6.94}$ & 25.42$_{1.06}$ \\
GPT2-medium & 81.00$_{2.16}$ & 93.68$_{0.86}$ & 80.00$_{0.86}$ & 44.53$_{2.46}$ & 26.11$_{1.41}$ & 16.97$_{3.97}$ & 21.74$_{3.20}$ & 46.81$_{1.74}$ & 17.39$_{0.00}$ & 27.36$_{3.52}$ & 62.67$_{0.47}$ & 75.00$_{2.92}$ & 63.96$_{1.27}$ & 58.33$_{2.36}$ & 38.93$_{1.60}$ & 70.02$_{10.50}$ & 23.54$_{1.93}$ \\
GPT2-small & 84.33$_{0.94}$ & 93.68$_{1.49}$ & 72.98$_{0.99}$ & 42.72$_{1.40}$ & 24.70$_{1.34}$ & 11.72$_{2.03}$ & 21.38$_{0.51}$ & 41.84$_{3.62}$ & 20.29$_{1.02}$ & 21.39$_{1.96}$ & 51.33$_{2.49}$ & 75.00$_{2.92}$ & 54.05$_{2.21}$ & 57.50$_{3.54}$ & 35.52$_{2.12}$ & 80.39$_{3.14}$ & 24.79$_{1.56}$ \\
SmoLLM2-1.7b & 95.67$_{1.70}$ & 99.65$_{0.50}$ & 87.37$_{1.49}$ & 59.69$_{0.97}$ & 35.65$_{3.36}$ & 41.68$_{1.50}$ & 31.88$_{2.56}$ & 41.13$_{1.00}$ & 31.88$_{1.02}$ & 31.84$_{5.18}$ & 74.67$_{1.70}$ & 90.48$_{1.68}$ & 72.07$_{2.55}$ & 76.67$_{4.25}$ & 46.07$_{1.71}$ & 50.04$_{3.63}$ & 24.58$_{3.96}$ \\
SmoLLM2-360m & 90.33$_{3.40}$ & 97.54$_{0.99}$ & 83.51$_{3.47}$ & 54.87$_{2.31}$ & 26.04$_{2.12}$ & 24.11$_{5.63}$ & 16.67$_{0.51}$ & 36.88$_{1.00}$ & 20.29$_{4.10}$ & 20.90$_{1.61}$ & 67.33$_{1.25}$ & 85.71$_{0.00}$ & 70.27$_{2.21}$ & 77.50$_{3.54}$ & 40.65$_{0.88}$ & 61.37$_{8.80}$ & 22.92$_{1.28}$ \\
\hline

\multicolumn{18}{c}{\cellcolor[HTML]{DDDDDD}\textbf{HELPFUL-HARMLESS}}\\

GPT2-large & 86.33$_{1.70}$ & 91.23$_{1.98}$ & 71.23$_{8.61}$ & 40.83$_{1.05}$ & 24.56$_{1.60}$ & 12.05$_{5.66}$ & 23.55$_{1.85}$ & 37.59$_{2.01}$ & 21.74$_{3.55}$ & 18.41$_{2.46}$ & 50.33$_{1.25}$ & 52.38$_{16.06}$ & 60.36$_{3.37}$ & 46.67$_{3.12}$ & 38.95$_{2.76}$ & 85.16$_{3.59}$ & 24.58$_{2.06}$ \\
GPT2-medium & 87.67$_{0.94}$ & 93.33$_{1.79}$ & 79.65$_{1.79}$ & 42.38$_{3.45}$ & 26.81$_{1.56}$ & 15.35$_{1.19}$ & 23.19$_{1.36}$ & 48.23$_{3.62}$ & 15.22$_{0.00}$ & 22.39$_{1.06}$ & 55.67$_{0.94}$ & 77.38$_{1.68}$ & 65.77$_{3.37}$ & 56.67$_{1.18}$ & 37.24$_{2.42}$ & 89.56$_{3.08}$ & 20.21$_{2.06}$ \\
GPT2-small & 84.33$_{0.94}$ & 90.18$_{1.31}$ & 68.07$_{3.47}$ & 45.39$_{3.30}$ & 23.72$_{1.01}$ & 10.64$_{0.91}$ & 22.46$_{2.23}$ & 41.84$_{3.62}$ & 20.29$_{3.69}$ & 20.15$_{4.27}$ & 50.33$_{2.05}$ & 64.29$_{5.83}$ & 51.35$_{2.21}$ & 52.50$_{2.04}$ & 37.76$_{1.89}$ & 85.53$_{9.23}$ & 23.54$_{0.59}$ \\
SmoLLM2-1.7b & 96.67$_{1.70}$ & 98.25$_{0.50}$ & 85.26$_{1.72}$ & 59.43$_{1.12}$ & 37.47$_{1.49}$ & 41.21$_{2.80}$ & 27.17$_{1.77}$ & 39.72$_{2.65}$ & 29.71$_{2.71}$ & 30.10$_{1.96}$ & 73.33$_{0.47}$ & 90.48$_{1.68}$ & 73.87$_{1.27}$ & 75.83$_{4.25}$ & 46.07$_{2.86}$ & 54.44$_{4.07}$ & 22.71$_{0.78}$ \\
SmoLLM2-360m & 87.00$_{2.16}$ & 96.84$_{0.00}$ & 77.89$_{2.98}$ & 53.66$_{0.80}$ & 23.86$_{1.76}$ & 21.82$_{1.87}$ & 19.93$_{0.51}$ & 45.39$_{3.62}$ & 20.29$_{1.02}$ & 16.42$_{3.71}$ & 58.33$_{2.62}$ & 76.19$_{3.37}$ & 73.87$_{3.37}$ & 66.67$_{5.14}$ & 40.02$_{0.68}$ & 64.65$_{12.49}$ & 23.54$_{2.30}$ \\
\toprule
\end{tabular}
\end{adjustbox}
\caption{Behavioural analysis on helpful-specific tasks. Column abbreviations: 
Alp‑E/H/L—AlpacaEval (Easy/Hard/Length),
Fact—Factuality,
LB—LLM Bar (Inst: adversarial instruction, Out: adversarial output,
Man: manual adversary, Nbr: neighboring adversary, Nat: natural),
MT—MT-Bench (Easy/Hard/Medium),
Prec‑IF—Precise Instruction Following.}
\label{tab:behavioural_analysis_helpful}
\end{table*}

\begin{table*}[!th]
\centering
\setlength{\tabcolsep}{3pt}
\renewcommand{\arraystretch}{1.1}
\begin{adjustbox}{max width=0.8\linewidth}
\begin{tabular}{l|cccccccc}
\hline
Model & \rotatebox{0}{DNA} &
\rotatebox{0}{RD} &
\rotatebox{0}{RO} &
\rotatebox{0}{Saf} &
\rotatebox{0}{SafR} &
\rotatebox{0}{SResp} &
\rotatebox{0}{XS-Ref} &
\rotatebox{0}{XS-Res}
% &Avg 
\\
\hline
\multicolumn{9}{c}{\cellcolor[HTML]{DDDDDD}\textbf{HARMLESS}}\\
\hline
GPT2-large & 86.76$_{1.04}$ & 100.00$_{0.00}$ & 100.00$_{0.00}$ & 66.89$_{1.01}$ & 99.26$_{0.36}$ & 8.63$_{1.00}$ & 99.35$_{0.53}$ & 3.20$_{0.98}$ \\

GPT2-medium & 86.03$_{1.59}$ & 100.00$_{0.00}$ & 100.00$_{0.00}$ & 66.89$_{1.75}$ & 98.87$_{0.11}$ & 10.97$_{0.53}$ & 99.57$_{0.61}$ & 6.00$_{0.86}$ \\

GPT2-small & 86.03$_{0.60}$ & 99.33$_{0.94}$ & 99.00$_{0.82}$ & 65.41$_{2.42}$ & 97.07$_{0.29}$ & 12.24$_{0.44}$ & 97.84$_{0.31}$ & 6.67$_{1.15}$ \\

SmoLLM2-1.7b & 91.67$_{1.39}$ & 99.67$_{0.47}$ & 100.00$_{0.00}$ & 76.52$_{2.12}$ & 99.77$_{0.10}$ & 8.21$_{0.61}$ & 99.78$_{0.31}$ & 4.40$_{1.50}$ \\

SmoLLM2-360m & 93.14$_{0.92}$ & 91.00$_{7.07}$ & 98.67$_{0.47}$ & 73.70$_{1.39}$ & 97.07$_{0.25}$ & 11.39$_{1.31}$ & 98.05$_{0.53}$ & 6.00$_{0.33}$ \\
\hline
\multicolumn{9}{c}{\cellcolor[HTML]{DDDDDD}\textbf{HELPFUL}}\\
\hline
GPT2-large & 10.78$_{3.52}$ & 0.00$_{0.00}$ & 13.33$_{4.03}$ & 8.15$_{2.09}$ & 4.23$_{1.41}$ & 94.48$_{2.38}$ & 11.04$_{6.93}$ & 91.47$_{4.16}$ \\

GPT2-medium & 8.82$_{0.00}$ & 0.00$_{0.00}$ & 17.67$_{1.70}$ & 6.52$_{0.42}$ & 5.05$_{0.67}$ & 92.43$_{0.72}$ & 7.58$_{1.86}$ & 93.07$_{0.68}$ \\

GPT2-small & 12.99$_{0.35}$ & 0.00$_{0.00}$ & 27.33$_{2.36}$ & 11.33$_{0.63}$ & 10.68$_{1.22}$ & 86.69$_{2.78}$ & 14.94$_{0.00}$ & 90.00$_{0.98}$ \\

SmoLLM2-1.7b & 13.48$_{1.51}$ & 0.00$_{0.00}$ & 16.33$_{4.50}$ & 16.37$_{1.03}$ & 4.58$_{0.69}$ & 97.45$_{0.46}$ & 12.77$_{2.21}$ & 95.60$_{0.57}$ \\

SmoLLM2-360m & 10.78$_{2.11}$ & 0.33$_{0.47}$ & 18.00$_{1.63}$ & 12.15$_{1.47}$ & 8.76$_{2.10}$ & 92.71$_{0.26}$ & 9.52$_{1.33}$ & 90.93$_{0.50}$ \\
\hline
\multicolumn{9}{c}{\cellcolor[HTML]{DDDDDD}\textbf{HELPFUL-HARMLESS}}\\
\hline
GPT2-large & 30.88$_{4.33}$ & 44.33$_{4.19}$ & 84.00$_{9.42}$ & 28.07$_{4.04}$ & 79.03$_{5.69}$ & 57.40$_{5.20}$ & 88.10$_{3.37}$ & 59.73$_{2.22}$ \\

GPT2-medium & 32.84$_{3.47}$ & 47.33$_{1.89}$ & 94.00$_{2.16}$ & 32.15$_{1.68}$ & 86.23$_{1.60}$ & 52.16$_{2.82}$ & 90.48$_{2.92}$ & 59.60$_{3.15}$ \\

GPT2-small & 30.88$_{1.20}$ & 37.67$_{2.49}$ & 86.33$_{1.70}$ & 27.56$_{1.42}$ & 79.97$_{0.28}$ & 52.80$_{3.61}$ & 80.52$_{2.31}$ & 62.53$_{2.10}$ \\

SmoLLM2-1.7b & 41.18$_{2.62}$ & 69.00$_{3.56}$ & 95.33$_{0.94}$ & 46.74$_{2.30}$ & 93.86$_{0.53}$ & 64.61$_{6.74}$ & 98.05$_{0.53}$ & 68.13$_{4.63}$ \\

SmoLLM2-360m & 30.88$_{1.20}$ & 35.67$_{2.05}$ & 90.33$_{3.40}$ & 37.70$_{1.59}$ & 88.42$_{1.01}$ & 62.49$_{1.77}$ & 89.61$_{1.40}$ & 67.33$_{3.64}$ \\
\hline
\end{tabular}
\end{adjustbox}
\caption{
Behavioural analysis in harmless-specific tasks.
Column abbreviations: DNA—Do Not Answer, 
RD—Refusals Dangerous, 
RO—Refusals Offensive, 
Saf—Safety, 
SafR—Safety Refuse, 
SResp—Safety Response, 
XS-Ref—XSTest Should Refuse, 
XS-Resp—XSTest Should Respond.
}
\label{tab:behavioural_analysis_harmless}
\end{table*}

\subsection{Reward model score}
\label{aba:RM_score}

\begin{figure}[!th]
    \centering
    \includegraphics[width=\linewidth]{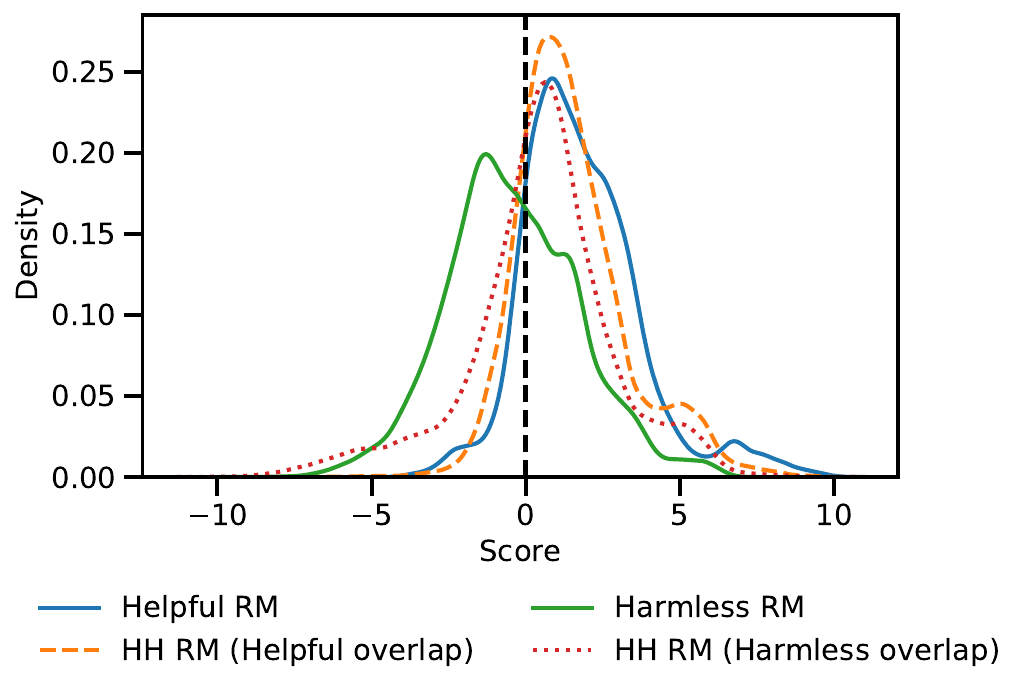}
    
    \caption{\textcolor{black}{Kernel density estimates of score between single-objective reward models and the mixed-objective reward model (HH).}}
    \label{fig:scoring_change}
% \vspace{-2mm}
\end{figure}

\subsection{Model-wise retention score}
\label{aba:mdoel_wise_transferance_score}

Table~\ref{tab:transferance_all} shows the retention score for all models along with its variance across the three seed values. Overall, we note low retention in all models across all tasks except in Alpaca Eval and refusal-based tasks.

\begin{table*}[h]
\centering
\small
\begin{tabular}{lcccccc}
\hline
\textbf{Task} & \textbf{S-1.7B} & \textbf{S-135M} & \textbf{S-360M} & \textbf{GPT2-S} & \textbf{GPT2-M} & \textbf{GPT2-L} \\
\hline
\multicolumn{7}{c}{\textbf{Helpfulness-specific Tasks}} \\
\hline
Alpaca Easy  & $0.969_{0.022}$ & $0.916_{0.022}$ & $0.915_{0.013}$ & $0.924_{0.019}$ & $0.945_{0.011}$ & $0.922_{0.007}$ \\
Alpaca Hard & $0.982_{0.005}$ & $0.954_{0.022}$ & $0.979_{0.000}$ & $0.929_{0.015}$ & $0.960_{0.027}$ & $0.943_{0.017}$ \\
Alpaca Length  & $0.867_{0.014}$ & $0.844_{0.008}$ & $0.842_{0.030}$ & $0.744_{0.045}$ & $0.823_{0.015}$ & $0.734_{0.089}$ \\
Chat  & $0.687_{0.023}$ & $0.601_{0.030}$ & $0.633_{0.022}$ & $0.520_{0.029}$ & $0.474_{0.029}$ & $0.448_{0.011}$ \\
Factuality  & $0.536_{0.024}$ & $0.405_{0.060}$ & $0.369_{0.045}$ & $0.344_{0.023}$ & $0.425_{0.026}$ & $0.322_{0.038}$ \\
Focus & $0.514_{0.019}$ & $0.216_{0.036}$ & $0.327_{0.058}$ & $0.133_{0.019}$ & $0.180_{0.022}$ & $0.128_{0.053}$ \\
LLM Bar (Natural)  & $0.761_{0.016}$ & $0.651_{0.020}$ & $0.619_{0.046}$ & $0.585_{0.007}$ & $0.628_{0.022}$ & $0.555_{0.012}$ \\
LLM Bar (Adver N.)  & $0.321_{0.029}$ & $0.239_{0.026}$ & $0.197_{0.037}$ & $0.221_{0.043}$ & $0.247_{0.026}$ & $0.199_{0.019}$ \\
LLM Bar (GPTInst) & $0.357_{0.033}$ & $0.242_{0.053}$ & $0.257_{0.035}$ & $0.242_{0.025}$ & $0.245_{0.022}$ & $0.236_{0.011}$ \\
LLM Bar (GPTOut) & $0.428_{0.016}$ & $0.531_{0.023}$ & $0.497_{0.043}$ & $0.460_{0.039}$ & $0.566_{0.059}$ & $0.382_{0.045}$ \\
LLM Bar (Manual) & $0.383_{0.014}$ & $0.234_{0.045}$ & $0.238_{0.025}$ & $0.184_{0.039}$ & $0.203_{0.012}$ & $0.221_{0.026}$ \\
MT Bench (Easy)& $0.927_{0.029}$ & $0.786_{0.020}$ & $0.810_{0.032}$ & $0.662_{0.050}$ & $0.837_{0.025}$ & $0.569_{0.168}$ \\
MT Bench (Med)  & $0.787_{0.046}$ & $0.612_{0.028}$ & $0.784_{0.049}$ & $0.606_{0.018}$ & $0.643_{0.009}$ & $0.491_{0.053}$ \\
MT Bench (Hard) & $0.819_{0.031}$ & $0.616_{0.057}$ & $0.779_{0.042}$ & $0.566_{0.029}$ & $0.676_{0.024}$ & $0.611_{0.017}$ \\
Math & $0.589_{0.048}$ & $0.569_{0.025}$ & $0.547_{0.027}$ & $0.537_{0.012}$ & $0.475_{0.041}$ & $0.495_{0.032}$ \\
Math PRM  & $0.690_{0.015}$ & $0.970_{0.023}$ & $0.785_{0.151}$ & $0.878_{0.096}$ & $0.906_{0.023}$ & $0.868_{0.042}$ \\
Precise IF & $0.369_{0.018}$ & $0.391_{0.043}$ & $0.422_{0.023}$ & $0.418_{0.026}$ & $0.397_{0.030}$ & $0.369_{0.029}$ \\
\hline
\multicolumn{7}{c}{\textbf{Harmless-specific Tasks}} \\
\hline
Do Not Answer & $0.431_{0.030}$ & $0.212_{0.028}$ & $0.321_{0.012}$ & $0.329_{0.010}$ & $0.342_{0.033}$ & $0.322_{0.039}$ \\
Refusals Dangerous  & $0.692_{0.039}$ & $0.180_{0.016}$ & $0.385_{0.008}$ & $0.379_{0.023}$ & $0.473_{0.019}$ & $0.443_{0.042}$ \\
Refusals Offensive & $0.953_{0.009}$ & $0.734_{0.041}$ & $0.903_{0.034}$ & $0.863_{0.017}$ & $0.940_{0.022}$ & $0.840_{0.094}$ \\
Safety& $0.529_{0.025}$ & $0.336_{0.018}$ & $0.445_{0.024}$ & $0.343_{0.027}$ & $0.405_{0.016}$ & $0.347_{0.049}$ \\
Safety Refuse & $0.939_{0.006}$ & $0.773_{0.015}$ & $0.891_{0.009}$ & $0.806_{0.001}$ & $0.865_{0.016}$ & $0.792_{0.057}$ \\
Safety Response & $0.649_{0.066}$ & $0.597_{0.047}$ & $0.639_{0.023}$ & $0.544_{0.041}$ & $0.532_{0.030}$ & $0.583_{0.058}$ \\
XStest Refuse  & $0.981_{0.005}$ & $0.702_{0.012}$ & $0.907_{0.006}$ & $0.804_{0.023}$ & $0.907_{0.030}$ & $0.881_{0.033}$ \\
XStest Respond & $0.698_{0.053}$ & $0.722_{0.070}$ & $0.716_{0.041}$ & $0.662_{0.017}$ & $0.622_{0.030}$ & $0.638_{0.028}$ \\
\hline
\end{tabular}
\caption{Retention score across models for the models covered in our analysis. Higher is better. Model abbreviations: S-1.7B (SmoLLM2-1.7B), S-360M (SmoLLM2-360M), S-135M (SmoLLM2-135M), GPT2-S (GPT-2 Small), GPT2-M (GPT-2 Medium), GPT2-L (GPT-2 Large).}
\label{tab:transferance_all}
\end{table*}

\begin{table*}[!th]
\centering
\begin{tabular}{l ccc}
\hline
Model & Helpful & Harmless & Helpful-Harmless\\
\hline
SmoLLM2-1.7b   & 0.00298 & 0.00448& 0.00373 \\
SmoLLM2-135m   & 0.00272 & 0.00480& 0.00274\\
SmoLLM2-360m   & 0.00275 & 0.00481& 0.00287\\
GPT2-small     & 0.00131 & 0.00244& 0.00159\\
GPT2-medium    & 0.00137 & 0.00258& 0.00195\\
GPT2-large     & 0.00117 & 0.00247& 0.00142\\
\hline
\end{tabular}
\caption{Global RMSD between chosen and rejected activations.}
\label{tab:rmsd_global}
\end{table*}

\begin{figure*}[!th]
    \centering
    \includegraphics[width=\textwidth]{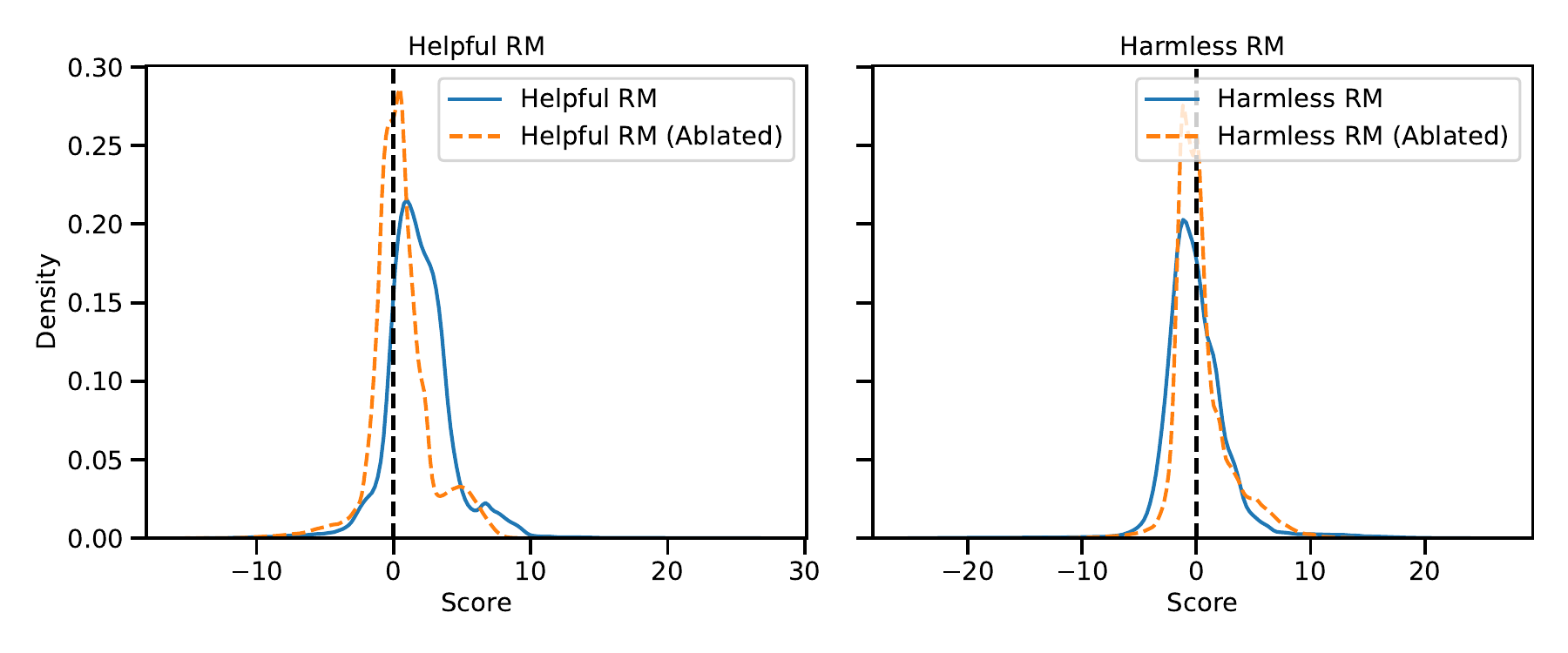}
    \caption{Score when single-objective mixed models are ablated using the objective-specific neurons.}
    \label{fig:scoring_abalated}
\end{figure*}

\subsection{Model-wise effect of shared neurons}
\label{aba:shared_neurons}
Table~\ref{tab:ablation_p10_Harmless_shapred} and Table~\ref{tab:ablation_p10_Helpful_shared} show the model-wise effect of ablating shared, harmless-only, and helpful-only neurons.
\begin{table*}[!th]
\centering
\setlength{\tabcolsep}{3pt}
\renewcommand{\arraystretch}{1.1}
\begin{adjustbox}{max width=\textwidth}
\begin{tabular}{l|cccccccc|c}
\hline
Model & \rotatebox{90}{Do Not Answer} & \rotatebox{90}{Refusals Dangerous} & \rotatebox{90}{Refusals Offensive} & \rotatebox{90}{Safety} & \rotatebox{90}{Safety Refuse} & \rotatebox{90}{Safety Response} & \rotatebox{90}{XStest Should Refuse} & \rotatebox{90}{XStest Should Respond} & Avg \\
\hline
\multicolumn{10}{c}{\cellcolor[HTML]{DDDDDD}\textbf{HELPFUL-HARMLESS}}\\
\textbf{Baseline} & 32.28 & 43.57 & 88.29 & 34.42 & 84.82 & 55.32 & 87.32 & 62.29 & 61.04 \\
\hdashline
$(-)\ shared$ & \cellcolor{blue!20}+6.62 & \cellcolor{red!20}-1.57 & \cellcolor{red!20}-23.05 & \cellcolor{red!20}-7.12 & \cellcolor{red!20}-21.27 & \cellcolor{red!20}-2.28 & \cellcolor{red!20}-25.48 & \cellcolor{red!20}-5.81 & \cellcolor{red!20}-10.00 \\
$(-)\ harmless\ only$ & \cellcolor{red!20}-4.20 & \cellcolor{red!20}-5.86 & \cellcolor{red!20}-3.81 & \cellcolor{red!20}-1.65 & \cellcolor{red!20}-4.77 & \cellcolor{blue!20}+9.65 & \cellcolor{red!20}-5.01 & \cellcolor{blue!20}+6.21 & \cellcolor{red!20}-1.18 \\
$(-)\ helpful\ only$ & \cellcolor{blue!20}+0.56 & \cellcolor{blue!20}+1.52 & \cellcolor{blue!20}+0.38 & \cellcolor{blue!20}+0.88 & \cellcolor{blue!20}+0.99 & \cellcolor{red!20}-1.93 & \cellcolor{red!20}-0.80 & \cellcolor{red!20}-1.62 & \cellcolor{red!20}-0.00 \\
\hline
\end{tabular}
\end{adjustbox}
\caption{Performance change relative to baseline ($\tau=0$) for $\tau=10$ on \textbf{Harmless} tasks.}
\label{tab:ablation_p10_Harmless_shapred}
\end{table*}

\begin{table*}[!th]
\centering
\setlength{\tabcolsep}{3pt}
\renewcommand{\arraystretch}{1.1}
\begin{adjustbox}{max width=\textwidth}
\begin{tabular}{l|ccccccccccccccccc|c}
\hline
Model & \rotatebox{90}{Alpaca Eval (Easy)} & \rotatebox{90}{Alpaca Eval (Hard)} & \rotatebox{90}{Alpaca Eval (Length)} & \rotatebox{90}{Chat} & \rotatebox{90}{Factuality} & \rotatebox{90}{Focus} & \rotatebox{90}{LLM Bar (Adver GPTInst)} & \rotatebox{90}{LLM Bar (Adver GPTOut)} & \rotatebox{90}{LLM Bar (Adver Manual)} & \rotatebox{90}{LLM Bar (Adver Neighbor)} & \rotatebox{90}{LLM Bar (Natural)} & \rotatebox{90}{MT Bench (Easy)} & \rotatebox{90}{MT Bench (Hard)} & \rotatebox{90}{MT Bench (Med)} & \rotatebox{90}{Math} & \rotatebox{90}{Math PRM} & \rotatebox{90}{Precise IF} & Avg \\
\hline
\multicolumn{19}{c}{\cellcolor[HTML]{DDDDDD}\textbf{HELPFUL-HARMLESS}}\\
\textbf{Baseline} & 88.62 & 94.24 & 77.69 & 49.35 & 28.05 & 19.12 & 22.77 & 43.26 & 20.50 & 21.68 & 58.38 & 73.81 & 64.35 & 61.43 & 40.65 & 71.55 & 22.50 & 50.47 \\
\hdashline
$(-)\ shared$ & \cellcolor{red!20}-18.67 & \cellcolor{red!20}-21.25 & \cellcolor{red!20}-17.64 & \cellcolor{red!20}-3.08 & \cellcolor{red!20}-1.94 & \cellcolor{red!20}-0.42 & \cellcolor{blue!20}+11.23 & \cellcolor{blue!20}+4.36 & \cellcolor{blue!20}+15.42 & \cellcolor{blue!20}+13.86 & \cellcolor{red!20}-7.86 & \cellcolor{red!20}-16.67 & \cellcolor{red!20}-11.20 & \cellcolor{red!20}-7.98 & \cellcolor{red!20}-2.55 & \cellcolor{red!20}-12.87 & \cellcolor{blue!20}+2.50 & \cellcolor{red!20}-4.40 \\
$(-)\ harmless\ only$ & \cellcolor{blue!20}+0.29 & \cellcolor{red!20}-0.25 & \cellcolor{red!20}-0.55 & \cellcolor{red!20}-0.52 & \cellcolor{red!20}-0.18 & \cellcolor{red!20}-1.05 & \cellcolor{red!20}-0.78 & \cellcolor{red!20}-1.11 & \cellcolor{red!20}-1.04 & \cellcolor{red!20}-0.53 & \cellcolor{red!20}-0.67 & \cellcolor{red!20}-2.04 & \cellcolor{red!20}-2.06 & \cellcolor{red!20}-0.24 & \cellcolor{blue!20}+0.01 & \cellcolor{red!20}-2.36 & \cellcolor{blue!20}+0.18 & \cellcolor{red!20}-0.76 \\
$(-)\ helpful\ only$ & \cellcolor{red!20}-1.71 & \cellcolor{red!20}-2.16 & \cellcolor{red!20}-1.05 & \cellcolor{red!20}-0.49 & \cellcolor{red!20}-1.01 & \cellcolor{red!20}-0.19 & \cellcolor{blue!20}+0.16 & \cellcolor{blue!20}+1.62 & \cellcolor{blue!20}+1.76 & \cellcolor{blue!20}+0.50 & \cellcolor{red!20}-1.43 & \cellcolor{red!20}-3.74 & \cellcolor{red!20}-2.19 & \cellcolor{blue!20}+0.24 & \cellcolor{red!20}-0.06 & \cellcolor{red!20}-4.77 & \cellcolor{blue!20}+0.39 & \cellcolor{red!20}-0.83 \\
\hline
\end{tabular}
\end{adjustbox}
\caption{Performance change relative to baseline ($\tau=0$) for $\tau=10$ on \textbf{Helpful} tasks.}
\label{tab:ablation_p10_Helpful_shared}
\end{table*}

\begin{table*}[!th]
\centering
\small
\setlength{\tabcolsep}{3pt}
\begin{tabular}{lcccc}
\hline
\textbf{Sub-task} & \textbf{Baseline} & \textbf{Helpful-only} & \textbf{Harmless-only} & \textbf{Shared} \\
\hline
Alpaca Eval (Easy)        & 0.933 & 0.923 & 0.926 & 0.875 \\
Alpaca Eval (Hard)        & 0.959 & 0.947 & 0.950 & 0.898 \\
Alpaca Eval (Length)      & 0.818 & 0.810 & 0.811 & 0.762 \\
Chat                      & 0.561 & 0.557 & 0.555 & 0.538 \\
Factuality                & 0.408 & 0.396 & 0.397 & 0.379 \\
Focus                     & 0.249 & 0.244 & 0.240 & 0.231 \\
LLM Bar (Adver GPTInst)   & 0.270 & 0.269 & 0.265 & 0.290 \\
LLM Bar (Adver GPTOut)    & 0.472 & 0.479 & 0.471 & 0.475 \\
LLM Bar (Adver Manual)    & 0.243 & 0.249 & 0.241 & 0.274 \\
LLM Bar (Adver Neighbor)  & 0.242 & 0.243 & 0.240 & 0.272 \\
LLM Bar (Natural)         & 0.643 & 0.634 & 0.634 & 0.608 \\
MT Bench (Easy)           & 0.783 & 0.766 & 0.764 & 0.720 \\
MT Bench (Hard)           & 0.678 & 0.664 & 0.660 & 0.629 \\
MT Bench (Med)            & 0.672 & 0.671 & 0.669 & 0.640 \\
Math                      & 0.533 & 0.526 & 0.526 & 0.501 \\
Math PRM                  & 0.772 & 0.744 & 0.744 & 0.708 \\
Precise IF                & 0.381 & 0.370 & 0.370 & 0.360 \\
\hline
Do Not Answer             & 0.338 & 0.340 & 0.324 & 0.341 \\
Refusals Dangerous        & 0.444 & 0.452 & 0.429 & 0.428 \\
Refusals Offensive        & 0.885 & 0.887 & 0.874 & 0.819 \\
Safety                    & 0.411 & 0.416 & 0.407 & 0.381 \\
Safety Refuse             & 0.852 & 0.857 & 0.839 & 0.789 \\
Safety Response           & 0.564 & 0.554 & 0.590 & 0.577 \\
XStest Should Refuse      & 0.876 & 0.872 & 0.856 & 0.797 \\
XStest Should Respond     & 0.658 & 0.649 & 0.674 & 0.651 \\
\hline
\end{tabular}
\caption{Retention across sub-tasks under different neuron ablation settings. \textbf{Baseline} denotes the original mixed objective model, \textbf{Helpful-only} and \textbf{Harmless-only} denote ablation of objective-specific neurons, and \textbf{Shared} denotes ablation of neurons common to both objectives.}
\label{tab:transfer_ablation_full}
\end{table*}
\subsection{Neural ablation comparison.}
\label{aba:neural_abalation_comparison}

\textcolor{black}{We compare zero ablation and mean ablation in Table~\ref{tab:hhrlhf_ablation_comparison}. Mean ablation consistently produces larger variance, whereas zero ablation yields more stable behaviour across models and reward-model types. Since our goal is to study the causal effect of reward-model representations while minimising unintended distributional changes, we use zero ablation in all subsequent experiments.}

\begin{table}[h] 
\centering 
\small 
\begin{tabular}{lcc} 
\toprule \textbf{Model} & \textbf{Mean Ablation} & \textbf{Zero ablation} \\

\hline
\multicolumn{3}{c}{\cellcolor[HTML]{DDDDDD}\textbf{HELPFUL-only}} \\ 
GPT2-large & $45.55_{11.17}$ & $64.95_{1.49}$ \\ 
GPT2-medium & $41.28_{19.55}$ & $50.92_{1.89}$ \\ 
GPT2-small & $65.31_{12.83}$ & $62.86_{2.16}$ \\ 
SmoLLM2-1.7b & $40.47_{19.49}$ & $50.38_{1.50}$ \\ 
SmoLLM2-135m & $34.23_{42.69}$ & $61.37_{1.75}$ \\ 
SmoLLM2-360m & $71.14_{15.39}$ & $57.66_{1.74}$ \\ \hline 
\multicolumn{3}{c}{\cellcolor[HTML]{DDDDDD}\textbf{HARMLESS-only}} \\
GPT2-large & $79.38_{9.89}$ & $67.98_{0.44}$ \\ 
GPT2-medium & $72.92_{9.54}$ & $52.47_{1.29}$ \\ 
GPT2-small & $74.18_{14.50}$ & $62.28_{1.28}$ \\ 
SmoLLM2-1.7b & $68.34_{12.47}$ & $52.34_{0.06}$ \\ 
SmoLLM2-135m & $69.28_{30.07}$ & $57.30_{3.48}$ \\ 
SmoLLM2-360m & $22.06_{30.40}$ & $49.21_{1.81}$ \\\hline
\multicolumn{3}{c}{\cellcolor[HTML]{DDDDDD}\textbf{HELPFUL-HARMLESS}}\\
GPT2-large & $55.61_{12.73}$ & $61.13_{3.09}$ \\ 
GPT2-medium & $27.43_{21.82}$ & $48.55_{1.02}$ \\ 
GPT2-small & $49.80_{14.43}$ & $57.22_{2.22}$ \\ 
SmoLLM2-1.7b & $71.65_{3.27}$ & $48.44_{1.20}$ \\ 
SmoLLM2-135m & $60.43_{41.04}$ & $52.91_{0.82}$ \\ 
SmoLLM2-360m & $6.69_{4.03}$ & $50.90_{4.63}$ \\ 
\bottomrule 
\end{tabular} 
\caption{\textcolor{black}{Accuracy (\%) after ablation on HH-RLHF reward-model directions. Values are reported as $\mathrm{accuracy}_{\mathrm{variance}}$.}} 
\label{tab:hhrlhf_ablation_comparison} 
\end{table}

\subsection{Model-wise effect of neuron ablation on behaviours}
\label{aba:abalation_behavious}

Table~\ref{tab:ablation_p10_Harmless} and~\ref{tab:ablation_p10_Helpful} show the effect of ablation on each of the RMs.

\begin{table}[t]
\centering
\setlength{\tabcolsep}{3pt}
\renewcommand{\arraystretch}{1.1}
\begin{adjustbox}{max width=0.5\textwidth}
\begin{tabular}{l|cccccccc|c}
\hline
Model & \rotatebox{90}{Do Not Answer} & \rotatebox{90}{Refusals Dangerous} & \rotatebox{90}{Refusals Offensive} & \rotatebox{90}{Safety} & \rotatebox{90}{Safety Refuse} & \rotatebox{90}{Safety Response} & \rotatebox{90}{XStest Should Refuse} & \rotatebox{90}{XStest Should Respond} & Avg \\
\hline
\multicolumn{10}{c}{\cellcolor[HTML]{DDDDDD}\textbf{HARMLESS}}\\
\multicolumn{10}{l}{\cellcolor[HTML]{EEEEEE}GPT2-large}\\
\textbf{Baseline} & 86.76 & 100.00 & 100.00 & 66.89 & 99.26 & 8.63 & 99.35 & 3.20 & 70.51 \\
\hdashline
$(-)\ top-\tau$ & \cellcolor{red!20}-0.74 & \cellcolor{blue!20}+0.00 & \cellcolor{blue!20}+0.00 & \cellcolor{red!20}-1.33 & \cellcolor{red!20}-0.23 & \cellcolor{blue!20}+2.83 & \cellcolor{red!20}-0.65 & \cellcolor{blue!20}+1.47 & \cellcolor{blue!20}+0.17 \\
\hline
\multicolumn{10}{l}{\cellcolor[HTML]{EEEEEE}GPT2-medium}\\
\textbf{Baseline} & 86.03 & 100.00 & 100.00 & 66.89 & 98.87 & 10.97 & 99.57 & 6.00 & 71.04 \\
\hdashline
$(-)\ top-\tau$ & \cellcolor{red!20}-38.24 & \cellcolor{red!20}-21.00 & \cellcolor{red!20}-41.67 & \cellcolor{red!20}-38.52 & \cellcolor{red!20}-48.51 & \cellcolor{blue!20}+44.44 & \cellcolor{red!20}-60.17 & \cellcolor{blue!20}+64.93 & \cellcolor{red!20}-17.34 \\
\hline
\multicolumn{10}{l}{\cellcolor[HTML]{EEEEEE}GPT2-small}\\
\textbf{Baseline} & 86.03 & 99.33 & 99.00 & 65.41 & 97.07 & 12.24 & 97.84 & 6.67 & 70.45 \\
\hdashline
$(-)\ top-\tau$ & \cellcolor{red!20}-4.41 & \cellcolor{red!20}-2.00 & \cellcolor{red!20}-4.67 & \cellcolor{red!20}-11.48 & \cellcolor{red!20}-3.17 & \cellcolor{blue!20}+9.06 & \cellcolor{red!20}-3.25 & \cellcolor{blue!20}+8.27 & \cellcolor{red!20}-1.46 \\
\hline
\multicolumn{10}{l}{\cellcolor[HTML]{EEEEEE}SmoLLM2-1.7b}\\
\textbf{Baseline} & 91.67 & 99.67 & 100.00 & 76.52 & 99.77 & 8.21 & 99.78 & 4.40 & 72.50 \\
\hdashline
$(-)\ top-\tau$ & \cellcolor{red!20}-36.76 & \cellcolor{red!20}-50.33 & \cellcolor{red!20}-46.67 & \cellcolor{red!20}-46.89 & \cellcolor{red!20}-39.55 & \cellcolor{blue!20}+33.83 & \cellcolor{red!20}-37.01 & \cellcolor{blue!20}+27.73 & \cellcolor{red!20}-24.46 \\
\hline
\multicolumn{10}{l}{\cellcolor[HTML]{EEEEEE}SmoLLM2-135m}\\
\textbf{Baseline} & 88.73 & 84.67 & 96.33 & 68.67 & 92.29 & 17.91 & 97.40 & 8.27 & 69.28 \\
\hdashline
$(-)\ top-\tau$ & \cellcolor{red!20}-13.97 & \cellcolor{red!20}-40.33 & \cellcolor{red!20}-28.67 & \cellcolor{red!20}-34.96 & \cellcolor{red!20}-20.50 & \cellcolor{blue!20}+23.00 & \cellcolor{red!20}-24.68 & \cellcolor{blue!20}+23.33 & \cellcolor{red!20}-14.60 \\
\hline
\multicolumn{10}{l}{\cellcolor[HTML]{EEEEEE}SmoLLM2-360m}\\
\textbf{Baseline} & 93.14 & 91.00 & 98.67 & 73.70 & 97.07 & 11.39 & 98.05 & 6.00 & 71.13 \\
\hdashline
$(-)\ top-\tau$ & \cellcolor{red!20}-38.48 & \cellcolor{red!20}-29.67 & \cellcolor{red!20}-42.67 & \cellcolor{red!20}-46.15 & \cellcolor{red!20}-31.85 & \cellcolor{blue!20}+34.25 & \cellcolor{red!20}-39.18 & \cellcolor{blue!20}+36.67 & \cellcolor{red!20}-19.63 \\
\hline
\multicolumn{10}{c}{\cellcolor[HTML]{DDDDDD}\textbf{HELPFUL}}\\
\multicolumn{10}{l}{\cellcolor[HTML]{EEEEEE}GPT2-large}\\
\textbf{Baseline} & 10.78 & 0.00 & 13.33 & 8.15 & 4.23 & 94.48 & 11.04 & 91.47 & 29.18 \\
\hdashline
$(-)\ top-\tau$ & \cellcolor{blue!20}+6.13 & \cellcolor{blue!20}+3.00 & \cellcolor{blue!20}+13.00 & \cellcolor{blue!20}+1.04 & \cellcolor{blue!20}+3.36 & \cellcolor{red!20}-6.72 & \cellcolor{blue!20}+3.68 & \cellcolor{red!20}-4.53 & \cellcolor{blue!20}+2.37 \\
\hline
\multicolumn{10}{l}{\cellcolor[HTML]{EEEEEE}GPT2-medium}\\
\textbf{Baseline} & 8.82 & 0.00 & 17.67 & 6.52 & 5.05 & 92.43 & 7.58 & 93.07 & 28.89 \\
\hdashline
$(-)\ top-\tau$ & \cellcolor{blue!20}+40.69 & \cellcolor{blue!20}+30.00 & \cellcolor{blue!20}+26.33 & \cellcolor{blue!20}+17.63 & \cellcolor{blue!20}+32.08 & \cellcolor{red!20}-36.59 & \cellcolor{blue!20}+31.39 & \cellcolor{red!20}-36.67 & \cellcolor{blue!20}+13.11 \\
\hline
\multicolumn{10}{l}{\cellcolor[HTML]{EEEEEE}GPT2-small}\\
\textbf{Baseline} & 12.99 & 0.00 & 27.33 & 11.33 & 10.68 & 86.69 & 14.94 & 90.00 & 31.75 \\
\hdashline
$(-)\ top-\tau$ & \cellcolor{blue!20}+5.39 & \cellcolor{blue!20}+2.67 & \cellcolor{red!20}-2.33 & \cellcolor{blue!20}+4.52 & \cellcolor{red!20}-2.03 & \cellcolor{blue!20}+0.42 & \cellcolor{blue!20}+6.28 & \cellcolor{red!20}-9.60 & \cellcolor{blue!20}+0.66 \\
\hline
\multicolumn{10}{l}{\cellcolor[HTML]{EEEEEE}SmoLLM2-1.7b}\\
\textbf{Baseline} & 13.48 & 0.00 & 16.33 & 16.37 & 4.58 & 97.45 & 12.77 & 95.60 & 32.07 \\
\hdashline
$(-)\ top-\tau$ & \cellcolor{blue!20}+12.75 & \cellcolor{blue!20}+19.00 & \cellcolor{blue!20}+32.33 & \cellcolor{blue!20}+4.44 & \cellcolor{blue!20}+26.25 & \cellcolor{red!20}-34.25 & \cellcolor{blue!20}+22.08 & \cellcolor{red!20}-31.73 & \cellcolor{blue!20}+6.36 \\
\hline
\multicolumn{10}{l}{\cellcolor[HTML]{EEEEEE}SmoLLM2-135m}\\
\textbf{Baseline} & 10.05 & 1.00 & 19.33 & 12.00 & 16.59 & 86.55 & 13.20 & 87.07 & 30.72 \\
\hdashline
$(-)\ top-\tau$ & \cellcolor{blue!20}+7.11 & \cellcolor{blue!20}+5.00 & \cellcolor{blue!20}+5.67 & \cellcolor{blue!20}+5.78 & \cellcolor{blue!20}+2.27 & \cellcolor{red!20}-15.00 & \cellcolor{blue!20}+9.96 & \cellcolor{red!20}-13.73 & \cellcolor{blue!20}+0.88 \\
\hline
\multicolumn{10}{l}{\cellcolor[HTML]{EEEEEE}SmoLLM2-360m}\\
\textbf{Baseline} & 10.78 & 0.33 & 18.00 & 12.15 & 8.76 & 92.71 & 9.52 & 90.93 & 30.40 \\
\hdashline
$(-)\ top-\tau$ & \cellcolor{blue!20}+25.49 & \cellcolor{blue!20}+23.33 & \cellcolor{blue!20}+23.33 & \cellcolor{blue!20}+11.56 & \cellcolor{blue!20}+26.29 & \cellcolor{red!20}-26.61 & \cellcolor{blue!20}+22.51 & \cellcolor{red!20}-31.20 & \cellcolor{blue!20}+9.34 \\
\hline
\multicolumn{10}{c}{\cellcolor[HTML]{DDDDDD}\textbf{HELPFUL-HARMLESS}}\\
\multicolumn{10}{l}{\cellcolor[HTML]{EEEEEE}GPT2-large}\\
\textbf{Baseline} & 30.88 & 44.33 & 84.00 & 28.07 & 79.03 & 57.40 & 88.10 & 59.73 & 58.94 \\
\hdashline
$(-)\ top-\tau$ & \cellcolor{red!20}-5.88 & \cellcolor{red!20}-8.67 & \cellcolor{red!20}-18.33 & \cellcolor{red!20}-5.26 & \cellcolor{red!20}-11.78 & \cellcolor{blue!20}+13.02 & \cellcolor{red!20}-18.61 & \cellcolor{blue!20}+12.27 & \cellcolor{red!20}-5.41 \\
\hline
\multicolumn{10}{l}{\cellcolor[HTML]{EEEEEE}GPT2-medium}\\
\textbf{Baseline} & 32.84 & 47.33 & 94.00 & 32.15 & 86.23 & 52.16 & 90.48 & 59.60 & 61.85 \\
\hdashline
$(-)\ top-\tau$ & \cellcolor{blue!20}+8.82 & \cellcolor{blue!20}+20.33 & \cellcolor{red!20}-47.33 & \cellcolor{red!20}-4.37 & \cellcolor{red!20}-38.15 & \cellcolor{red!20}-2.34 & \cellcolor{red!20}-48.05 & \cellcolor{blue!20}+3.07 & \cellcolor{red!20}-13.50 \\
\hline
\multicolumn{10}{l}{\cellcolor[HTML]{EEEEEE}GPT2-small}\\
\textbf{Baseline} & 30.88 & 37.67 & 86.33 & 27.56 & 79.97 & 52.80 & 80.52 & 62.53 & 57.28 \\
\hdashline
$(-)\ top-\tau$ & \cellcolor{blue!20}+2.94 & \cellcolor{red!20}-11.67 & \cellcolor{red!20}-14.67 & \cellcolor{red!20}-0.22 & \cellcolor{red!20}-20.19 & \cellcolor{blue!20}+17.98 & \cellcolor{red!20}-16.88 & \cellcolor{blue!20}+8.67 & \cellcolor{red!20}-4.26 \\
\hline
\multicolumn{10}{l}{\cellcolor[HTML]{EEEEEE}SmoLLM2-1.7b}\\
\textbf{Baseline} & 41.18 & 69.00 & 95.33 & 46.74 & 93.86 & 64.61 & 98.05 & 68.13 & 72.11 \\
\hdashline
$(-)\ top-\tau$ & \cellcolor{blue!20}+2.94 & \cellcolor{red!20}-30.33 & \cellcolor{red!20}-48.33 & \cellcolor{red!20}-19.48 & \cellcolor{red!20}-42.14 & \cellcolor{red!20}-14.30 & \cellcolor{red!20}-50.00 & \cellcolor{red!20}-19.33 & \cellcolor{red!20}-27.62 \\
\hline
\multicolumn{10}{l}{\cellcolor[HTML]{EEEEEE}SmoLLM2-135m}\\
\textbf{Baseline} & 20.34 & 15.33 & 72.00 & 28.22 & 76.25 & 57.61 & 69.70 & 66.13 & 50.70 \\
\hdashline
$(-)\ top-\tau$ & \cellcolor{blue!20}+4.17 & \cellcolor{blue!20}+7.00 & \cellcolor{red!20}-24.33 & \cellcolor{red!20}-11.04 & \cellcolor{red!20}-30.67 & \cellcolor{blue!20}+8.56 & \cellcolor{red!20}-28.14 & \cellcolor{red!20}-1.87 & \cellcolor{red!20}-9.54 \\
\hline
\multicolumn{10}{l}{\cellcolor[HTML]{EEEEEE}SmoLLM2-360m}\\
\textbf{Baseline} & 30.88 & 35.67 & 90.33 & 37.70 & 88.42 & 62.49 & 89.61 & 67.33 & 62.81 \\
\hdashline
$(-)\ top-\tau$ & \cellcolor{blue!20}+22.06 & \cellcolor{blue!20}+1.00 & \cellcolor{red!20}-47.33 & \cellcolor{red!20}-14.81 & \cellcolor{red!20}-31.18 & \cellcolor{red!20}-19.18 & \cellcolor{red!20}-40.48 & \cellcolor{red!20}-20.00 & \cellcolor{red!20}-18.74 \\
\hline
\end{tabular}
\end{adjustbox}
\caption{Performance change relative to baseline ($\tau=0$) for $\tau=10$ on \textbf{Harmless} tasks.}
\label{tab:ablation_p10_Harmless}
\end{table}

\begin{table*}[t]
\centering
\setlength{\tabcolsep}{3pt}
\renewcommand{\arraystretch}{1.1}
\begin{adjustbox}{max width=\textwidth}
\begin{tabular}{l|ccccccccccccccccc|c}
\hline
Model & \rotatebox{90}{Alpaca Eval (Easy)} & \rotatebox{90}{Alpaca Eval (Hard)} & \rotatebox{90}{Alpaca Eval (Length)} & \rotatebox{90}{Chat} & \rotatebox{90}{Factuality} & \rotatebox{90}{Focus} & \rotatebox{90}{LLM Bar (Adver GPTInst)} & \rotatebox{90}{LLM Bar (Adver GPTOut)} & \rotatebox{90}{LLM Bar (Adver Manual)} & \rotatebox{90}{LLM Bar (Adver Neighbor)} & \rotatebox{90}{LLM Bar (Natural)} & \rotatebox{90}{MT Bench (Easy)} & \rotatebox{90}{MT Bench (Hard)} & \rotatebox{90}{MT Bench (Med)} & \rotatebox{90}{Math} & \rotatebox{90}{Math PRM} & \rotatebox{90}{Precise IF} & Avg \\
\hline
\multicolumn{19}{c}{\cellcolor[HTML]{DDDDDD}\textbf{HARMLESS}}\\
\multicolumn{19}{l}{\cellcolor[HTML]{EEEEEE}GPT2-large}\\
\textbf{Baseline} & 8.00 & 10.53 & 34.74 & 53.49 & 30.81 & 45.66 & 76.45 & 65.96 & 67.39 & 75.37 & 45.00 & 41.67 & 52.25 & 53.33 & 35.88 & 74.27 & 25.21 & 46.82 \\
\hdashline
$(-)\ top-\tau$ & \cellcolor{blue!20}+0.33 & \cellcolor{blue!20}+1.05 & \cellcolor{blue!20}+0.70 & \cellcolor{blue!20}+3.62 & \cellcolor{red!20}-1.19 & \cellcolor{red!20}-0.20 & \cellcolor{red!20}-1.81 & \cellcolor{red!20}-2.13 & \cellcolor{blue!20}+5.07 & \cellcolor{red!20}-0.50 & \cellcolor{blue!20}+3.67 & \cellcolor{blue!20}+5.95 & \cellcolor{red!20}-2.70 & \cellcolor{red!20}-5.83 & \cellcolor{blue!20}+0.59 & \cellcolor{blue!20}+6.71 & \cellcolor{blue!20}+1.04 & \cellcolor{blue!20}+0.85 \\
\hline
\multicolumn{19}{l}{\cellcolor[HTML]{EEEEEE}GPT2-medium}\\
\textbf{Baseline} & 6.67 & 8.77 & 40.35 & 53.49 & 30.67 & 46.33 & 74.28 & 60.28 & 68.84 & 76.87 & 46.33 & 36.90 & 50.45 & 49.17 & 35.34 & 76.36 & 22.50 & 46.09 \\
\hdashline
$(-)\ top-\tau$ & \cellcolor{blue!20}+50.33 & \cellcolor{blue!20}+41.05 & \cellcolor{blue!20}+3.16 & \cellcolor{red!20}-4.13 & \cellcolor{blue!20}+3.16 & \cellcolor{red!20}-25.19 & \cellcolor{red!20}-30.07 & \cellcolor{red!20}-5.67 & \cellcolor{red!20}-16.67 & \cellcolor{red!20}-33.33 & \cellcolor{blue!20}+3.33 & \cellcolor{blue!20}+4.76 & \cellcolor{blue!20}+0.90 & \cellcolor{red!20}-0.83 & \cellcolor{blue!20}+2.35 & \cellcolor{red!20}-52.95 & \cellcolor{blue!20}+1.88 & \cellcolor{red!20}-3.41 \\
\hline
\multicolumn{19}{l}{\cellcolor[HTML]{EEEEEE}GPT2-small}\\
\textbf{Baseline} & 7.33 & 7.72 & 39.30 & 52.02 & 32.14 & 45.39 & 69.57 & 65.96 & 68.12 & 78.86 & 41.33 & 46.43 & 56.76 & 57.50 & 35.37 & 64.28 & 23.75 & 46.58 \\
\hdashline
$(-)\ top-\tau$ & \cellcolor{blue!20}+9.00 & \cellcolor{blue!20}+14.39 & \cellcolor{blue!20}+3.16 & \cellcolor{red!20}-0.69 & \cellcolor{red!20}-5.96 & \cellcolor{blue!20}+1.08 & \cellcolor{red!20}-3.26 & \cellcolor{red!20}-7.80 & \cellcolor{red!20}-8.70 & \cellcolor{red!20}-7.21 & \cellcolor{blue!20}+3.67 & \cellcolor{blue!20}+15.48 & \cellcolor{red!20}-10.81 & \cellcolor{red!20}-8.33 & \cellcolor{blue!20}+1.56 & \cellcolor{blue!20}+13.42 & \cellcolor{red!20}-0.83 & \cellcolor{blue!20}+0.48 \\
\hline
\multicolumn{19}{l}{\cellcolor[HTML]{EEEEEE}Model Average}\\
\multicolumn{19}{l}{\cellcolor[HTML]{EEEEEE}SmoLLM2-1.7b}\\
\textbf{Baseline} & 21.67 & 17.89 & 37.19 & 48.32 & 33.82 & 30.91 & 54.35 & 65.25 & 58.70 & 72.89 & 43.67 & 38.10 & 37.84 & 41.67 & 34.58 & 32.44 & 30.83 & 41.18 \\
\hdashline
$(-)\ top-\tau$ & \cellcolor{blue!20}+26.00 & \cellcolor{blue!20}+23.86 & \cellcolor{blue!20}+8.42 & \cellcolor{red!20}-3.79 & \cellcolor{red!20}-6.67 & \cellcolor{red!20}-3.57 & \cellcolor{red!20}-7.25 & \cellcolor{red!20}-21.28 & \cellcolor{red!20}-9.42 & \cellcolor{red!20}-6.72 & \cellcolor{blue!20}+3.33 & \cellcolor{blue!20}+5.95 & \cellcolor{blue!20}+8.11 & \cellcolor{blue!20}+5.00 & \cellcolor{blue!20}+1.98 & \cellcolor{blue!20}+10.96 & \cellcolor{red!20}-8.54 & \cellcolor{blue!20}+1.55 \\
\hline
\multicolumn{19}{l}{\cellcolor[HTML]{EEEEEE}SmoLLM2-135m}\\
\textbf{Baseline} & 28.67 & 24.91 & 37.19 & 47.46 & 35.23 & 29.16 & 57.61 & 61.70 & 61.59 & 70.15 & 43.00 & 25.00 & 54.05 & 45.00 & 35.79 & 18.79 & 27.50 & 41.34 \\
\hdashline
$(-)\ top-\tau$ & \cellcolor{blue!20}+10.00 & \cellcolor{blue!20}+15.44 & \cellcolor{blue!20}+8.42 & \cellcolor{blue!20}+1.81 & \cellcolor{red!20}-9.61 & \cellcolor{blue!20}+0.61 & \cellcolor{red!20}-0.36 & \cellcolor{red!20}-9.93 & \cellcolor{red!20}-4.35 & \cellcolor{red!20}-12.94 & \cellcolor{blue!20}+3.67 & \cellcolor{blue!20}+11.90 & \cellcolor{blue!20}+1.80 & \cellcolor{red!20}-4.17 & \cellcolor{red!20}-1.18 & \cellcolor{blue!20}+26.70 & \cellcolor{red!20}-1.67 & \cellcolor{blue!20}+2.13 \\
\hline
\multicolumn{19}{l}{\cellcolor[HTML]{EEEEEE}SmoLLM2-360m}\\
\textbf{Baseline} & 34.67 & 25.26 & 36.49 & 46.08 & 32.49 & 26.60 & 55.80 & 63.12 & 68.12 & 71.39 & 47.33 & 29.76 & 45.05 & 38.33 & 33.57 & 18.94 & 27.92 & 41.23 \\
\hdashline
$(-)\ top-\tau$ & \cellcolor{blue!20}+9.67 & \cellcolor{blue!20}+15.09 & \cellcolor{blue!20}+14.39 & \cellcolor{blue!20}+3.70 & \cellcolor{red!20}-8.07 & \cellcolor{blue!20}+0.20 & \cellcolor{red!20}-2.17 & \cellcolor{red!20}-1.42 & \cellcolor{red!20}-18.12 & \cellcolor{red!20}-10.95 & \cellcolor{blue!20}+3.00 & \cellcolor{blue!20}+21.43 & \cellcolor{blue!20}+11.71 & \cellcolor{blue!20}+15.83 & \cellcolor{blue!20}+3.66 & \cellcolor{blue!20}+22.60 & \cellcolor{blue!20}+0.83 & \cellcolor{blue!20}+4.79 \\
\hline
\multicolumn{19}{c}{\cellcolor[HTML]{DDDDDD}\textbf{HELPFUL}}\\
\multicolumn{19}{l}{\cellcolor[HTML]{EEEEEE}GPT2-large}\\
\textbf{Baseline} & 81.00 & 91.23 & 73.33 & 45.48 & 23.23 & 16.50 & 28.26 & 43.26 & 18.12 & 20.15 & 53.67 & 67.86 & 59.46 & 53.33 & 38.99 & 77.85 & 25.42 & 48.07 \\
\hdashline
$(-)\ top-\tau$ & \cellcolor{red!20}-23.33 & \cellcolor{red!20}-18.25 & \cellcolor{red!20}-12.98 & \cellcolor{red!20}-1.12 & \cellcolor{red!20}-4.91 & \cellcolor{blue!20}+5.19 & \cellcolor{blue!20}+11.96 & \cellcolor{blue!20}+4.26 & \cellcolor{blue!20}+5.07 & \cellcolor{blue!20}+1.00 & \cellcolor{red!20}-3.33 & \cellcolor{red!20}-9.52 & \cellcolor{red!20}-0.90 & \cellcolor{red!20}-9.17 & \cellcolor{red!20}-2.30 & \cellcolor{red!20}-24.38 & \cellcolor{red!20}-1.46 & \cellcolor{red!20}-4.95 \\
\hline
\multicolumn{19}{l}{\cellcolor[HTML]{EEEEEE}GPT2-medium}\\
\textbf{Baseline} & 81.00 & 93.68 & 80.00 & 44.53 & 26.11 & 16.97 & 21.74 & 46.81 & 17.39 & 27.36 & 62.67 & 75.00 & 63.96 & 58.33 & 38.93 & 70.02 & 23.54 & 49.89 \\
\hdashline
$(-)\ top-\tau$ & \cellcolor{red!20}-27.67 & \cellcolor{red!20}-31.23 & \cellcolor{red!20}-34.74 & \cellcolor{blue!20}+5.00 & \cellcolor{red!20}-0.77 & \cellcolor{blue!20}+8.28 & \cellcolor{blue!20}+25.36 & \cellcolor{blue!20}+0.71 & \cellcolor{blue!20}+27.54 & \cellcolor{blue!20}+14.93 & \cellcolor{red!20}-13.33 & \cellcolor{red!20}-26.19 & \cellcolor{red!20}-11.71 & \cellcolor{red!20}-8.33 & \cellcolor{red!20}-3.83 & \cellcolor{red!20}-28.86 & \cellcolor{blue!20}+7.50 & \cellcolor{red!20}-5.73 \\
\hline
\multicolumn{19}{l}{\cellcolor[HTML]{EEEEEE}GPT2-small}\\
\textbf{Baseline} & 84.33 & 93.68 & 72.98 & 42.72 & 24.70 & 11.72 & 21.38 & 41.84 & 20.29 & 21.39 & 51.33 & 75.00 & 54.05 & 57.50 & 35.52 & 80.39 & 24.79 & 47.86 \\
\hdashline
$(-)\ top-\tau$ & \cellcolor{red!20}-7.33 & \cellcolor{red!20}-11.23 & \cellcolor{red!20}-6.32 & \cellcolor{red!20}-0.69 & \cellcolor{red!20}-6.88 & \cellcolor{blue!20}+2.49 & \cellcolor{blue!20}+7.97 & \cellcolor{red!20}-0.71 & \cellcolor{blue!20}+5.80 & \cellcolor{blue!20}+6.72 & \cellcolor{red!20}-5.33 & \cellcolor{red!20}-11.90 & \cellcolor{red!20}-0.90 & \cellcolor{red!20}-3.33 & \cellcolor{blue!20}+2.32 & \cellcolor{red!20}-11.19 & \cellcolor{red!20}-2.29 & \cellcolor{red!20}-2.52 \\
\hline

\multicolumn{19}{l}{\cellcolor[HTML]{EEEEEE}SmoLLM2-1.7b}\\
\textbf{Baseline} & 95.67 & 99.65 & 87.37 & 59.69 & 35.65 & 41.68 & 31.88 & 41.13 & 31.88 & 31.84 & 74.67 & 90.48 & 72.07 & 76.67 & 46.07 & 50.04 & 24.58 & 58.30 \\
\hdashline
$(-)\ top-\tau$ & \cellcolor{red!20}-15.67 & \cellcolor{red!20}-19.65 & \cellcolor{red!20}-37.19 & \cellcolor{red!20}-17.14 & \cellcolor{red!20}-12.49 & \cellcolor{red!20}-23.23 & \cellcolor{red!20}-7.25 & \cellcolor{blue!20}+7.80 & \cellcolor{blue!20}+4.35 & \cellcolor{red!20}-1.24 & \cellcolor{red!20}-24.00 & \cellcolor{red!20}-42.86 & \cellcolor{red!20}-19.82 & \cellcolor{red!20}-25.83 & \cellcolor{red!20}-10.13 & \cellcolor{red!20}-19.31 & \cellcolor{blue!20}+0.42 & \cellcolor{red!20}-15.49 \\
\hline
\multicolumn{19}{l}{\cellcolor[HTML]{EEEEEE}SmoLLM2-135m}\\
\textbf{Baseline} & 92.00 & 97.89 & 77.19 & 50.13 & 30.18 & 12.86 & 16.67 & 34.04 & 15.94 & 19.15 & 58.67 & 80.95 & 63.96 & 68.33 & 38.77 & 89.63 & 18.54 & 50.88 \\
\hdashline
$(-)\ top-\tau$ & \cellcolor{red!20}-16.00 & \cellcolor{red!20}-16.14 & \cellcolor{red!20}-19.65 & \cellcolor{red!20}-6.03 & \cellcolor{red!20}-9.54 & \cellcolor{blue!20}+4.65 & \cellcolor{blue!20}+13.77 & \cellcolor{blue!20}+7.09 & \cellcolor{blue!20}+10.14 & \cellcolor{blue!20}+6.72 & \cellcolor{red!20}-4.33 & \cellcolor{red!20}-16.67 & \cellcolor{red!20}-11.71 & \cellcolor{red!20}-9.17 & \cellcolor{red!20}-0.70 & \cellcolor{red!20}-29.83 & \cellcolor{blue!20}+4.58 & \cellcolor{red!20}-5.46 \\
\hline
\multicolumn{19}{l}{\cellcolor[HTML]{EEEEEE}SmoLLM2-360m}\\
\textbf{Baseline} & 90.33 & 97.54 & 83.51 & 54.87 & 26.04 & 24.11 & 16.67 & 36.88 & 20.29 & 20.90 & 67.33 & 85.71 & 70.27 & 77.50 & 40.65 & 61.37 & 22.92 & 52.76 \\
\hdashline
$(-)\ top-\tau$ & \cellcolor{red!20}-20.33 & \cellcolor{red!20}-32.98 & \cellcolor{red!20}-34.04 & \cellcolor{red!20}-7.41 & \cellcolor{red!20}-0.21 & \cellcolor{red!20}-6.33 & \cellcolor{blue!20}+19.57 & \cellcolor{blue!20}+14.18 & \cellcolor{blue!20}+11.59 & \cellcolor{blue!20}+17.91 & \cellcolor{red!20}-18.67 & \cellcolor{red!20}-38.10 & \cellcolor{red!20}-14.41 & \cellcolor{red!20}-29.17 & \cellcolor{red!20}-4.42 & \cellcolor{red!20}-16.41 & \cellcolor{red!20}-4.79 & \cellcolor{red!20}-9.65 \\
\hline
\multicolumn{19}{c}{\cellcolor[HTML]{DDDDDD}\textbf{HELPFUL-HARMLESS}}\\
\multicolumn{19}{l}{\cellcolor[HTML]{EEEEEE}GPT2-large}\\
\textbf{Baseline} & 86.33 & 91.23 & 71.23 & 40.83 & 24.56 & 12.05 & 23.55 & 37.59 & 21.74 & 18.41 & 50.33 & 52.38 & 60.36 & 46.67 & 38.95 & 85.16 & 24.58 & 46.23 \\
\hdashline
$(-)\ top-\tau$ & \cellcolor{red!20}-0.33 & \cellcolor{red!20}-0.70 & \cellcolor{red!20}-6.67 & \cellcolor{red!20}-1.64 & \cellcolor{red!20}-3.86 & \cellcolor{blue!20}+1.89 & \cellcolor{blue!20}+0.00 & \cellcolor{blue!20}+4.26 & \cellcolor{blue!20}+2.17 & \cellcolor{red!20}-1.49 & \cellcolor{blue!20}+3.67 & \cellcolor{blue!20}+3.57 & \cellcolor{blue!20}+2.70 & \cellcolor{blue!20}+7.50 & \cellcolor{red!20}-1.96 & \cellcolor{red!20}-31.39 & \cellcolor{red!20}-1.04 & \cellcolor{red!20}-1.37 \\
\hline
\multicolumn{19}{l}{\cellcolor[HTML]{EEEEEE}GPT2-medium}\\
\textbf{Baseline} & 87.67 & 93.33 & 79.65 & 42.38 & 26.81 & 15.35 & 23.19 & 48.23 & 15.22 & 22.39 & 55.67 & 77.38 & 65.77 & 56.67 & 37.24 & 89.56 & 20.21 & 50.39 \\
\hdashline
$(-)\ top-\tau$ & \cellcolor{red!20}-29.00 & \cellcolor{red!20}-34.39 & \cellcolor{red!20}-40.70 & \cellcolor{blue!20}+5.00 & \cellcolor{blue!20}+6.25 & \cellcolor{blue!20}+11.85 & \cellcolor{blue!20}+18.84 & \cellcolor{blue!20}+0.00 & \cellcolor{blue!20}+29.71 & \cellcolor{blue!20}+18.91 & \cellcolor{red!20}-1.33 & \cellcolor{red!20}-41.67 & \cellcolor{red!20}-13.51 & \cellcolor{red!20}-7.50 & \cellcolor{red!20}-0.49 & \cellcolor{red!20}-51.83 & \cellcolor{blue!20}+10.42 & \cellcolor{red!20}-7.03 \\
\hline
\multicolumn{19}{l}{\cellcolor[HTML]{EEEEEE}GPT2-small}\\
\textbf{Baseline} & 84.33 & 90.18 & 68.07 & 45.39 & 23.72 & 10.64 & 22.46 & 41.84 & 20.29 & 20.15 & 50.33 & 64.29 & 51.35 & 52.50 & 37.76 & 85.53 & 23.54 & 46.61 \\
\hdashline
$(-)\ top-\tau$ & \cellcolor{red!20}-14.67 & \cellcolor{red!20}-21.40 & \cellcolor{red!20}-5.26 & \cellcolor{red!20}-1.21 & \cellcolor{red!20}-3.09 & \cellcolor{blue!20}+4.58 & \cellcolor{blue!20}+14.13 & \cellcolor{blue!20}+8.51 & \cellcolor{blue!20}+8.70 & \cellcolor{blue!20}+10.70 & \cellcolor{red!20}-7.00 & \cellcolor{red!20}-10.71 & \cellcolor{red!20}-1.80 & \cellcolor{blue!20}+0.83 & \cellcolor{blue!20}+0.69 & \cellcolor{red!20}-17.45 & \cellcolor{blue!20}+0.62 & \cellcolor{red!20}-1.99 \\
\hline
\multicolumn{19}{l}{\cellcolor[HTML]{EEEEEE}SmoLLM2-1.7b}\\
\textbf{Baseline} & 96.67 & 98.25 & 85.26 & 59.43 & 37.47 & 41.21 & 27.17 & 39.72 & 29.71 & 30.10 & 73.33 & 90.48 & 73.87 & 75.83 & 46.07 & 54.44 & 22.71 & 57.75 \\
\hdashline
$(-)\ top-\tau$ & \cellcolor{red!20}-40.33 & \cellcolor{red!20}-42.46 & \cellcolor{red!20}-34.74 & \cellcolor{red!20}-13.09 & \cellcolor{red!20}-13.19 & \cellcolor{red!20}-16.57 & \cellcolor{blue!20}+13.04 & \cellcolor{blue!20}+7.80 & \cellcolor{blue!20}+16.67 & \cellcolor{blue!20}+19.90 & \cellcolor{red!20}-20.67 & \cellcolor{red!20}-42.86 & \cellcolor{red!20}-25.23 & \cellcolor{red!20}-21.67 & \cellcolor{red!20}-9.29 & \cellcolor{red!20}-13.57 & \cellcolor{blue!20}+4.79 & \cellcolor{red!20}-13.61 \\
\hline
\multicolumn{19}{l}{\cellcolor[HTML]{EEEEEE}SmoLLM2-135m}\\
\textbf{Baseline} & 88.67 & 95.09 & 77.89 & 50.30 & 26.67 & 11.99 & 17.75 & 47.52 & 18.12 & 20.65 & 55.00 & 71.43 & 59.46 & 55.83 & 40.66 & 94.26 & 21.25 & 50.15 \\
\hdashline
$(-)\ top-\tau$ & \cellcolor{red!20}-30.67 & \cellcolor{red!20}-30.53 & \cellcolor{red!20}-22.11 & \cellcolor{red!20}-1.81 & \cellcolor{red!20}-2.60 & \cellcolor{blue!20}+5.25 & \cellcolor{blue!20}+19.57 & \cellcolor{blue!20}+2.13 & \cellcolor{blue!20}+16.67 & \cellcolor{blue!20}+19.90 & \cellcolor{red!20}-7.33 & \cellcolor{red!20}-22.62 & \cellcolor{red!20}-5.41 & \cellcolor{red!20}-5.00 & \cellcolor{red!20}-2.59 & \cellcolor{red!20}-25.43 & \cellcolor{blue!20}+2.71 & \cellcolor{red!20}-5.29 \\
\hline
\multicolumn{19}{l}{\cellcolor[HTML]{EEEEEE}SmoLLM2-360m}\\
\textbf{Baseline} & 87.00 & 96.84 & 77.89 & 53.66 & 23.86 & 21.82 & 19.93 & 45.39 & 20.29 & 16.42 & 58.33 & 76.19 & 73.87 & 66.67 & 40.02 & 64.65 & 23.54 & 50.96 \\
\hdashline
$(-)\ top-\tau$ & \cellcolor{red!20}-34.33 & \cellcolor{red!20}-44.56 & \cellcolor{red!20}-28.77 & \cellcolor{red!20}-2.93 & \cellcolor{blue!20}+3.86 & \cellcolor{blue!20}+0.94 & \cellcolor{blue!20}+23.91 & \cellcolor{blue!20}+4.96 & \cellcolor{blue!20}+28.26 & \cellcolor{blue!20}+35.07 & \cellcolor{red!20}-6.00 & \cellcolor{red!20}-28.57 & \cellcolor{red!20}-19.82 & \cellcolor{red!20}-24.17 & \cellcolor{red!20}-2.09 & \cellcolor{red!20}-8.43 & \cellcolor{blue!20}+1.25 & \cellcolor{red!20}-5.97 \\
\hline
\end{tabular}
\end{adjustbox}
\caption{Performance change relative to baseline ($\tau=0$) for $\tau=10$ on \textbf{Helpful} tasks.}
\label{tab:ablation_p10_Helpful}
\end{table*}

\subsection{Objective-specific neurons distribution}
\label{aba:abalation_neuron_distribution}

\begin{figure}[!t]
\begin{center}
\includegraphics[width=\linewidth]{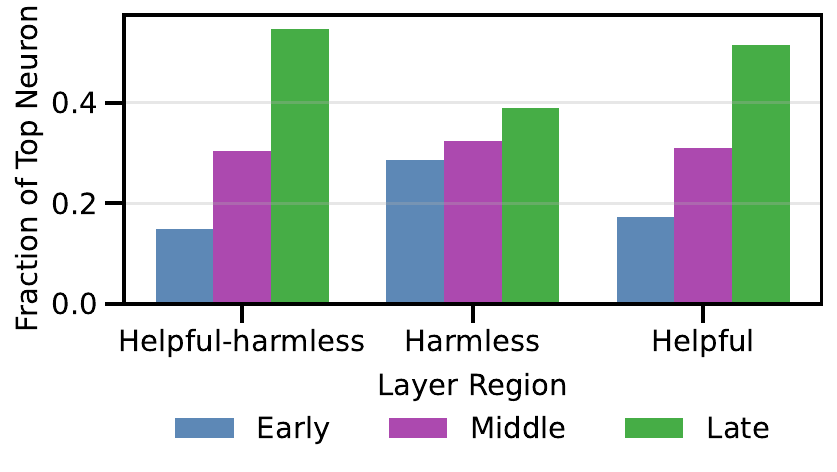}
\caption{\textcolor{black}{\textbf{Distribution of objective-specific neurons across layers.} We note that on average the neurons reside in the top model layers.}}
\label{fig:layer_distribution}
\end{center}
% \vspace{-5mm}
\end{figure}
\textcolor{black}{\textbf{Where are the neurons located?} Figure~\ref{fig:layer_distribution} analyses the depth-wise location of the selected objective-specific neurons by dividing the LLM component of RM into three equal buckets: early, middle, and late layers. We observe that, for all three RM types, \textit{objective-specific neurons are concentrated in the late layers}. In an LLM, these layers are shown to capture higher-level semantic and task-relevant features, whereas earlier layers capture surface-level linguistic structure~\cite{song2026demystifying,skean2025layer}.}\looseness=-1

\subsection{Chosen vs. rejected processing}
\label{sec:aba_chosen_vs_rej}

\textcolor{black}{\textbf{Chosen vs. rejected. }As described in \S\ref{sec:background}, an RM is trained to assign higher scores to responses chosen vs. rejected by a human. This raises a natural question: are different neurons processing chosen vs. rejected inputs? To examine this, given a preference triple $(p, y^+, y^-)\in \mathcal{D}$ and its associated neural activations $a_{j}^{l}$; we compute the root mean square difference (RMSD) between activations on the two responses as: \[
\text{RMSD} =
\sqrt{
\frac{1}{n}
\sum_{j=1}^{n}
\left(
a_{j}^{l}(p \oplus y^{+})
-
a_{j}^{l}(p \oplus y^{-})
\right)^2
}.
\]
We obtain RMSD values of $2.6 \times 10^{-3}$, $8.2\times 10^{-3}$, and $2.3\times10^{-3}$ for helpful, harmless and helpful-harmless RMs (ref. App. Table~\ref{tab:rmsd_global} for model-wise scores). These differences are small, indicating that \textit{mostly the same neurons are involved in processing both chosen and rejected responses}.}

\subsection{Task-wise direction agreement score}
\label{aba:task_wise_direction_chage}

\textcolor{black}{The  direction agreement score across all models fro each task is represented in Table~\ref{tab:patching_task_wise_helpful} and~\ref{tab:patching_task_wise_harmlessness}} 

\begin{table*}[!th] \centering \setlength{\tabcolsep}{3pt} \renewcommand{\arraystretch}{1.1} \begin{adjustbox}{max width=\textwidth} \begin{tabular}{l|cccccccc} \hline Patching type & \rotatebox{90}{Do Not Answer} & \rotatebox{90}{Refusals Dangerous} & \rotatebox{90}{Refusals Offensive} & \rotatebox{90}{Safety} & \rotatebox{90}{Safety Refuse} & \rotatebox{90}{Safety Response} & \rotatebox{90}{XStest Should Refuse} & \rotatebox{90}{XStest Should Respond} \\ \hline \multicolumn{9}{c}{\cellcolor[HTML]{DDDDDD}\textbf{HELPFUL-HARMLESS}}\\ \hline Shared Neurons& -0.67 & 0.00 & -0.33 & -1.00 & -0.67 & -0.67 & -0.33 & -0.33  \\ Unique Neurons& 0.67 & 0.67 & 0.33 & -0.33 & 1.00 & 0.67 & 0.00 & 0.00 \\ Random (Unique) & 0.00 & 0.00 & 0.00 & 0.00 & 0.00 & 0.00 & 0.00 & 0.00  \\ Random (shared)& 0.33 & 0.67 & 0.33 & 0.00 & 0.67 & 0.33 & 0.33 & 0.33\\ \hline \end{tabular} \end{adjustbox} \caption{Direction change score of the average performance change of models on \textbf{Harmlessness} tasks.} \label{tab:patching_task_wise_harmlessness} \end{table*}

\begin{table*}[!th] \centering \setlength{\tabcolsep}{3pt} \renewcommand{\arraystretch}{1.1} \begin{adjustbox}{max width=\textwidth} \begin{tabular}{l|ccccccccccccccccc} \hline Patching type & \rotatebox{90}{Alpaca Eval (Easy)} & \rotatebox{90}{Alpaca Eval (Hard)} & \rotatebox{90}{Alpaca Eval (Length)} & \rotatebox{90}{Chat} & \rotatebox{90}{Factuality} & \rotatebox{90}{Focus} & \rotatebox{90}{LLM Bar (Adver GPTInst)} & \rotatebox{90}{LLM Bar (Adver GPTOut)} & \rotatebox{90}{LLM Bar (Adver Manual)} & \rotatebox{90}{LLM Bar (Adver Neighbor)} & \rotatebox{90}{LLM Bar (Natural)} & \rotatebox{90}{MT Bench (Easy)} & \rotatebox{90}{MT Bench (Hard)} & \rotatebox{90}{MT Bench (Med)} & \rotatebox{90}{Math} & \rotatebox{90}{Math PRM} & \rotatebox{90}{Precise IF} \\ \hline \multicolumn{18}{c}{\cellcolor[HTML]{DDDDDD}\textbf{HELPFUL-HARMLESS}}\\ \hline Shared Neurons& -0.33 & -0.67 & -0.33 & 0.00 & 0.33 & -0.67 & 0.00 & 0.33 & -0.33 & 0.33 & 0.17 & -0.50 & 0.33 & -0.33 & 1.00 & -0.33 & 0.33  \\ Unique Neurons& -0.33 & 0.17 & -0.33 & 0.67 & 0.33 & 0.00 & -0.17 & 0.00 & 0.00 & 0.00 & 0.00 & 0.33 & 0.17 & -0.50 & 0.33 & -0.33 & 0.17 \\ Random (Unique) & 0.00 & 0.00 & 0.00 & 0.00 & 0.00 & 0.17 & 0.00 & 0.00 & 0.00 & 0.00 & 0.00 & 0.00 & 0.00 & 0.00 & 0.17 & 0.00 & 0.00 \\ Random (shared)& 0.00 & 0.00 & -0.33 & 0.33 & 0.33 & 0.00 & 0.17 & -0.33 & 0.33 & 0.33 & 0.33 & 0.50 & -0.50 & -0.67 & 1.00 & 0.00 & 0.00 \\ \hline \end{tabular} \end{adjustbox} \caption{Direction change score of the average performance change of models on \textbf{Helpful} tasks.} \label{tab:patching_task_wise_helpful} \end{table*}

\end{document}